\documentclass[nohyperref]{article}

\usepackage[accepted]{icml2023}

\usepackage[utf8]{inputenc} 
\usepackage[T1]{fontenc}    
\usepackage{hyperref}       
\usepackage{xcolor}
\usepackage{url}            
\usepackage{booktabs}       
\usepackage{nicefrac}       
\usepackage{microtype}      
\usepackage{soul}   
\usepackage{mathtools}
\usepackage{tabularx}
\usepackage{graphicx}
\usepackage[normalem]{ulem}
\usepackage{xspace}
\usepackage{multirow}
\usepackage{amsmath}
\usepackage{amssymb}
\usepackage{amsthm}
\usepackage{amsfonts}       
\usepackage{wrapfig}
\usepackage{adjustbox}
\usepackage{multicol}
\usepackage{mwe}
\usepackage[inline, shortlabels]{enumitem}
\setlist[enumerate]{nosep}
\usepackage[capitalize,noabbrev]{cleveref}
\usepackage{bm}
\usepackage{bbm}
\usepackage{array}
\usepackage{rotating}       
\usepackage{verbatim}
\usepackage{caption}
\usepackage{subcaption}    
\usepackage{tcolorbox}

\DeclareGraphicsExtensions{.pdf, .png}

\newcommand{\jihye}[1]{{\color{blue}[Jihye: #1]}}
\newcommand{\jayaram}[1]{\textcolor{red}{Jayaram: #1}}

\newcommand{\ryan}[1]{\textcolor{brown}{Ryan: #1}}

\theoremstyle{plain}
\newtheorem{theorem}{Theorem}

\theoremstyle{definition}
\newtheorem{definition}[theorem]{Definition}

\theoremstyle{remark}

\newcommand{\argmax}{\operatornamewithlimits{arg\!\max}}
\newcommand{\ben}{\begin{enumerate}}
\newcommand{\een}{\end{enumerate}}
\newcommand{\beq}{\begin{equation}}
\newcommand{\eeq}{\end{equation}}
\newcommand{\beqa}{\begin{eqnarray}}
\newcommand{\eeqa}{\end{eqnarray}}
\newcommand{\bit}{\begin{itemize}}
\newcommand{\eit}{\end{itemize}}
\newcommand{\btab}{\begin{tabular}}
\newcommand{\etab}{\end{tabular}}

\newcommand{\noprint}[1]{}

\newcommand{\mypara}[1]{\noindent\textbf{#1}}

\def\bibfont{\small}

\def \ie {{\em i.e.},~}

\def \eg {{\em e.g.},~}

\def \etal {{\em et al.}}

\def \reals {\mathbb{R}}

\def \expec {\mathop{\mathbb{E}}}

\def \indicator {\mathbbm{1}}

\newcommand{\inputs}{\mathcal{X}} 
\newcommand{\outputs}{\mathcal{Y}} 
\newcommand{\simplex}{\Delta} 
\newcommand{\Pin}{P_{\textrm{in}}}
\newcommand{\Pout}{P_{\textrm{out}}}
\newcommand{\Dtr}{D^{\textrm{tr}}}
\newcommand{\Dintr}{D_{\textrm{in}}^{\textrm{tr}}}
\newcommand{\Douttr}{D_{\textrm{out}}^{\textrm{tr}}}
\newcommand{\Dinte}{D_{\textrm{in}}^{\textrm{te}}}
\newcommand{\Doutte}{D_{\textrm{out}}^{\textrm{te}}}
\newcommand{\Dval}{D^{\textrm{val}}}
\newcommand{\Dinval}{D_{\textrm{in}}^{\textrm{val}}}
\newcommand{\Doutval}{D_{\textrm{out}}^{\textrm{val}}}
\newcommand{\Nintr}{N_{\textrm{in}}^{\textrm{tr}}}
\newcommand{\Nouttr}{N_{\textrm{out}}^{\textrm{tr}}}

\newcommand{\recphi}{\widehat{\bfphi}_{\bfg, \bfC}}
\newcommand{\VC}{\mathbf{v}^{}_{\mathbf{C}}}
\newcommand{\TVC}{\widetilde{\mathbf{v}}^{}_{\mathbf{C}}}
\newcommand{\fcon}{\mathbf{f}^{\textrm{con}}}
\newcommand{\Dcon}{\mathcal{D}^{\textrm{con}}_\gamma}
\newcommand{\Scon}{S^{\textrm{con}}}

\newcommand{\expnum}[2]{{#1}\mathrm{e}{#2}}

\def \mysum {\displaystyle\sum\limits}

\def \bfphi {\bm{\phi}}

\def \bfmu {\bm{\mu}}

\def \bfc {\mathbf{c}}

\def \bff {\mathbf{f}}
\def \bfg {\mathbf{g}}
\def \bfh {\mathbf{h}}

\def \bfv {\mathbf{v}}

\def \bfx {\mathbf{x}}

\def \bfC {\mathbf{C}}

\def \bfS {\mathbf{S}}

\def \calD {\mathcal{D}}

\def \calX {\mathcal{X}}

\def \calZ {\mathcal{Z}}

\usepackage{natbib}

\graphicspath{{figures/}}

\icmltitlerunning{Concept-based Explanations for Out-of-Distribution Detectors}

\begin{document}
\twocolumn[
\icmltitle{Concept-based Explanations for Out-of-Distribution Detectors}

\icmlsetsymbol{equal}{*}
\begin{icmlauthorlist}
\icmlauthor{Jihye Choi}{uw}
\icmlauthor{Jayaram Raghuram}{uw}
\icmlauthor{Ryan Feng}{um}
\icmlauthor{Jiefeng Chen}{uw}
\icmlauthor{Somesh Jha}{uw}
\icmlauthor{Atul Prakash}{um}
\end{icmlauthorlist}

\icmlaffiliation{uw}{University of Wisconsin - Madison}
\icmlaffiliation{um}{University of Michigan}

\icmlcorrespondingauthor{Jihye Choi}{jihye@cs.wisc.edu}

\icmlkeywords{Machine Learning, OOD detection, out-of-distribution, concept vectors, explainability, detection completeness}

\vskip 0.3in
]




\printAffiliationsAndNotice{}  

\begin{abstract}
Out-of-distribution (OOD) detection plays a crucial role in ensuring the safe deployment of deep neural network (DNN) classifiers. 
While a myriad of methods have focused on improving the performance of OOD detectors, a critical gap remains in interpreting their decisions.
We help bridge this gap by providing explanations for OOD detectors based on learned high-level concepts.
We first propose two new metrics for assessing the effectiveness of a particular set of concepts for explaining OOD detectors: 1) \textit{detection completeness} -- which quantifies the sufficiency of concepts for explaining an OOD detector's decisions, and 2) \textit{concept separability} -- which captures the distributional separation between in-distribution and OOD data in the concept space.
Based on these metrics, we propose an unsupervised framework for learning a set of concepts that satisfy the desired properties of high detection completeness and concept separability, and demonstrate its effectiveness in providing concept-based explanations for diverse off-the-shelf OOD detectors.
We also show how to identify prominent concepts contributing to the detection results, and provide further reasoning about their decisions.


\end{abstract}

\section{Introduction}

It is well known that machine learning (ML) models can yield uncertain and unreliable predictions on out-of-distribution (OOD) inputs, \ie inputs from outside the training distribution~\citep{amodei2016AISafety,goodfellow2015explaining,hendrycks2021many}.
A common line of defense in this situation is to augment the ML model (\eg a DNN classifier) with a detector that can identify and flag such inputs as OOD~\citep{hendrycks2016msp, liang2018ODIN}. 
The ML model can then abstain from making predictions on such inputs~\citep{tax2008growing, geifman2019selectivenet}.
In many application domains (\eg medical imaging), it is important to understand both the model's prediction as well the reason for abstaining from prediction on certain inputs (\ie the OOD detector's decisions).
Moreover, abstaining from prediction can often have a practical cost, \eg due to service denial or the need for manual intervention~\citep{markoff2016self, mozannar2020consistent}.

Detecting OOD inputs has received significant attention in the literature, and a number of methods exist that achieve strong detection performance on semantic distribution shifts~\cite{yang2021survey, yang2022openood}.
Much of the focus in learning OOD detectors has been on improving their detection performance~\citep{hendrycks2018OE, liu2020energy, mohseni2020self, lin2021MOOD, chen2021atom, sun2021react, cao2022deep}. 
However, the problem of explaining the decisions of an OOD detector and the related problem of designing inherently-interpretable detectors remain largely unexplored (we focus on the former problem). 
A potential approach could be to run an existing explanation method for DNN classifiers with in-distribution (ID) and OOD data separately, and then inspect the difference between the generated explanations. However, it is unclear whether an explanation method that is effective for ID class predictions will also be effective for OOD detection. For instance, feature attributions, the most popular type of explanation \citep{sundararajan2017IG,ribeiro2016LIME}, may not capture visual differences in the generated explanations between ID and OOD inputs~\citep{adebayo2020debugging}.
Moreover, their explanations based on pixel-level activations may not provide the most intuitive form of explanations for humans.

\begin{figure}[t]
     \centering
     \begin{subfigure}[b]{\columnwidth}
         \centering
         \includegraphics[width=\textwidth]{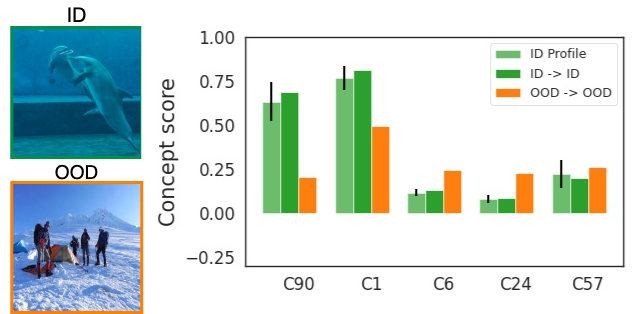}
         \caption{\small Correct detection: top dolphin image is correctly detected as ID (dark-green bar), and the bottom image is correctly detected as OOD (orange bar).}
         \label{fig:fig1a}
     \end{subfigure}
     \\
     \vspace{1mm}
     \begin{subfigure}[b]{\columnwidth}
         \centering
         \includegraphics[width=\textwidth]{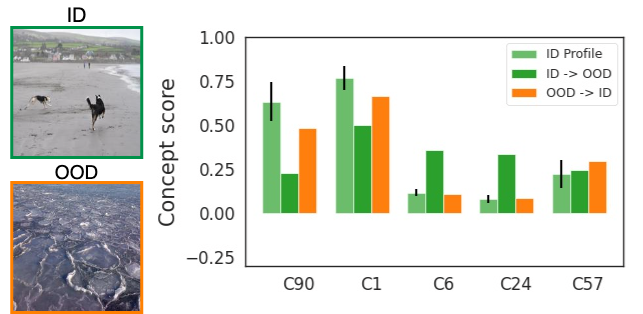}
         \caption{\small Wrong detection: top ID image is detected as OOD (dark-green bar), and the bottom OOD image is detected as ID (orange bar).}
         \label{fig:fig1b}
     \end{subfigure}
     \\
     \vspace{1mm}
     \begin{subfigure}[b]{\columnwidth}
         \hspace{1.5mm} \includegraphics[width=\textwidth]{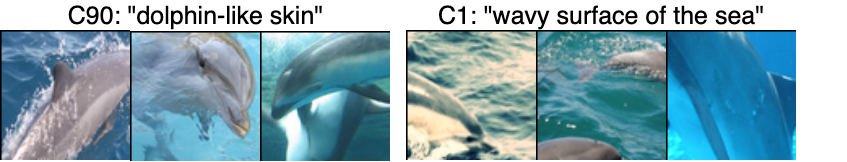}
         \caption{\small Visualization of top-2 important concepts.}
     \end{subfigure}
     \caption{\small \textbf{Our concept-based explanation for the Energy OOD detector~\citep{liu2020energy}.} All input images are classified as ``Dolphin'' but detected differently. On the x-axis of bar graphs, we present the top-5 important concepts that describe the detector's behavior given images classified as ``Dolphin''.
    ID profile (light-green) shows the average concept score pattern for ID images predicted as ``Dolphin''. We expect ID inputs predicted into this class to have a similar concept score pattern.}
\vspace{-5mm}
\label{fig:expl-ours-dolphin}
\end{figure}

This paper addresses the above research gap by proposing the first method (to our knowledge) for interpreting the decisions of an OOD detector in a human-understandable way.
We build upon recent advances in concept-based explanations for DNN classifiers~\citep{ghorbani2019ace,zhou2018interpretable,bouchacourt2019educe,yeh2020completeness}, which offer the benefit of providing explanations in terms of high-level {\em concepts} for classification tasks.
We focus on extending this explanation framework to the problem of OOD detection.
As a concrete example, consider Fig.~\ref{fig:expl-ours-dolphin} which illustrates our concept-based explanations given inputs which are all classified as the class ``Dolphin'' by a DNN classifier, but detected as either ID or OOD by an OOD detector.
Our method identifies that the concepts such as C90 ``dolphin-like skin'' and C1 ``wavy surface of the sea'' are key concepts to understand the OOD detector's decisions to tell apart ID and OOD images predicted as ``Dolphin''.
The user can verify that these concepts are aligned with human intuition and the OOD detector relies on them for making decisions.
We also confirm that the OOD detector predicts a certain input as ID when its concept-score patterns are similar to that of normal ID Dolphin images. Likewise, the detector predicts an input as OOD when its concept-score patterns are very different from that of ID inputs from the same class.
Our explanations can help a user analyze if an incorrect detection (as in Fig.~\ref{fig:fig1b}) is an understandable mistake or a misbehavior of the OOD detector, evaluate the reliability of the OOD detector, and decide upon its adoption in practice.




We aim to design a general interpretability framework that is applicable across a wide range of black-box OOD detectors.
Accordingly, a research question we ask is: {\em without relying on the internal mechanism of an OOD detector, can we identify a good set of concepts that are appropriate for understanding why the OOD detector predicts a certain input to be ID\,/\,OOD?} 
A key contribution of this paper is to show that this can be done unsupervised, without any additional human annotations for interpretation.

In summary, we make the following contributions:
\begin{itemize} [leftmargin=*, topsep=1pt, noitemsep]
\item We motivate and propose new metrics to quantify the effectiveness of concept-based explanations for a black-box OOD detector, namely \textit{detection completeness} and \textit{concept separability} (Sections \ref{sec:two_worlds}, \ref{sec:completeness_score}, and \ref{sec:separability_score}).  
\item We propose a concept-learning objective with suitable regularization terms that, given an OOD detector for a DNN classifier, learns a set of concepts with high detection completeness and concept separability (Section~\ref{sec:concept_learning});
\item By treating the OOD detector as a black-box, we show that our approach can be applied to explain a variety of existing OOD detection methods.
We also provide empirical evidence that concepts learned for classifiers cannot be directly used to explain OOD detectors, whereas concepts learned by our method are effective for explaining both the classifier and OOD detector (Section~\ref{sec:eval-concept}).
\item By identifying prominent concepts that contribute to an OOD detector's decisions via a modified Shapley value importance score based on the detection completeness, we demonstrate how the discovered concepts can be used to interpret the OOD detector (Section~\ref{sec:expt_concept_based_explanations}).
\end{itemize}

\mypara{Related Work. }  In the literature of OOD detection, recent studies have designed various scoring functions based on the representation from the final or penultimate layers \citep{liang2018ODIN, devries2018learning}, or a combination of different internal layers of a DNN classifier~\citep{lee2018mahalanobis, lin2021MOOD, raghuram2021JTLA}.
A recent survey on generalized OOD detection can be found in Yang et al.~\citep{yang2021survey}. Our work aims to provide post-hoc explanations applicable to a wide range of black-box OOD detectors without modifying their internals.
Among different interpretability approaches, concept-based explanation~\citep{koh2020concept-bottleneck, SENN} has gained popularity as it is designed to be better-aligned with human reasoning \citep{armstrong1983human-concepts,tenenbaum1999concept-learning} and intuition~\citep{ghorbani2019ace,zhou2018interpretable,bouchacourt2019educe,yeh2020completeness}. There have been limited attempts to assess the use of concept-based explanations under data distribution changes such as adversarial manipulation~\citep{kim2018tcav} or spurious correlations~\citep{adebayo2020debugging}.
However, designing concept-based explanations for OOD detection requires further exploration and is the focus of our work.

\section{Problem Setup and Background}
\mypara{Notations.}
Let $\inputs \subseteq \reals^{a_0 \times b_0 \times d_0}$ denote the space of inputs $\bfx$, where $d_0$ is the number of channels and $a_0$ and $b_0$ are the image size along each channel.
Let $\outputs := \{1, \cdots, L\}$ denote the space of output class labels $y$.
Let $\simplex_L$ denote the set of all probabilities over $\outputs$ (the simplex in $L$-dimensions).
We assume that natural inputs to the DNN classifier are sampled from an unknown probability distribution $\Pin$ over the space $\calX \times \outputs$.  
The  compact notation $[n]$ denotes $\{1, \cdots, n\}$ for a positive integer $n$.
Boldface symbols are used to denote both vectors and tensors.
$\langle\bfx, \bfx^\prime\rangle$ denotes the standard inner-product between a pair of vectors. 
The indicator function $\indicator[c]$ takes value $1$ ($0$) when the condition $c$ is true (false).

\mypara{ID and OOD Datasets.}
Consider a labeled ID training dataset $\Dintr = \{(\bfx_i, y_i), ~i = 1, \cdots, \Nintr\}$ sampled from the distribution $\Pin$.
We assume the availability of an unlabeled training dataset $\Douttr = \{\widetilde{\bfx}_i, ~i = 1, \cdots, \Nouttr\}$ from a different distribution, referred to as the {\em auxiliary OOD dataset}.
Similarly, we define the ID test dataset (from $\Pin$) as $\Dinte$, and the OOD test dataset as $\Doutte$.
Note that the auxiliary OOD dataset $\Dintr$ and the test OOD dataset $\Doutte$ are from different distributions.
All the OOD datasets are unlabeled since their label space is usually different from $\outputs$. 

\mypara{OOD Detector.}
The goal of an OOD detector is to determine if a test input to the classifier is ID (\ie from the distribution $\Pin$); otherwise the input is declared to be OOD~\citep{yang2021survey}.
Given a trained classifier $\bff : \calX \mapsto \Delta_L$, the decision function of an OOD detector can be generally defined as $\,\calD_\gamma(\bfx, \bff) \,=\, \indicator[S(\bfx, \bff) \geq \gamma]$, where $S(\bfx, \bff) \in \reals$ is the score function of the detector for an input $\bfx$ and $\gamma$ is the threshold.
%
%
We follow the convention that larger scores correspond to ID inputs, and the detector outputs of $1$ and $0$ correspond to ID and OOD respectively.
We assume the availability a pre-trained DNN classifier and a paired OOD detector that is 
trained to detect inputs for the classifier.

\subsection{Projection Into Concept Space}
\label{sec:concept_projection}

\begin{figure}[ht]
\centering
\includegraphics[scale=0.17]{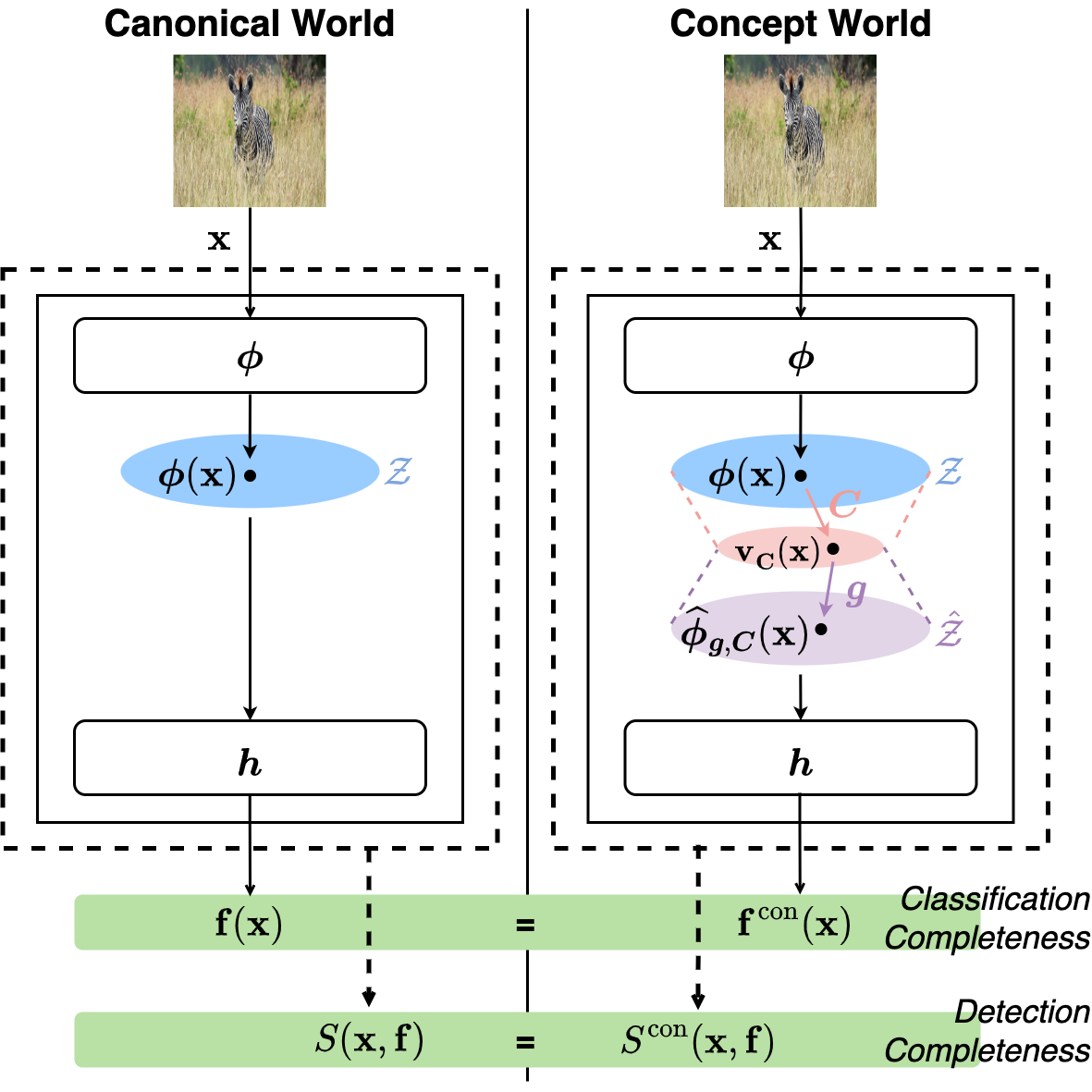} 
\caption{
\textbf{Our two-world view of the classifier and OOD detector.} In the canonical world, both the classifier and OOD detector are unmodified. In the concept world, the layer representation $\bfphi(\bfx)$ is projected into the space spanned by the concept vectors and then reconstructed via the non-linear mapping $\bfg$. The classifier and OOD detector in the concept world are based on this reconstructed layer representation. Given the same input, the outputs from the DNN classifier and OOD detector in both the worlds should be very close to each other (characterized by \textit{Classification Completeness} and \textit{Detection Completeness}, respectively).
}
\label{fig:detection-completeness}
\vspace{-0.19in}
\end{figure}
%

Consider a pre-trained DNN classifier $\bff : \mathcal{X} \mapsto \simplex_L$ that maps an input $\bfx$ to its corresponding predicted class probabilities. 
Without loss of generality, we can partition the DNN at a convolutional layer $\ell$ into two parts, \ie $\bff = \bfh \circ \bfphi$ where: 1) $\bfphi : \inputs \mapsto \calZ := \reals^{a_\ell b_\ell \times d_\ell}\,$ is the first half of $\bff$ that maps an input $\bfx$ to the intermediate feature representation~\footnote{We flatten the first two dimensions of the feature representation, thus changing an $a_\ell \times b_\ell \times d_\ell$ tensor to an $a_\ell b_\ell \times d_\ell$ matrix, where $a_\ell$ and $b_\ell$ are the filter size and $d_\ell$ is the number of channels.} $\bfphi(\bfx)$, and 2) $\bfh : \calZ \mapsto \simplex_L$ is the second half of $\bff$ that maps $\bfphi(\bfx)$ to the predicted class probabilities $\bfh(\bfphi(\bfx))$.
We denote the predicted probability of a class $y$ by $\,f_y(\bfx) = h_y(\bfphi(\bfx))$, and the prediction of the classifier by $\,\widehat{y}(\bfx) = \argmax_{y} f_y(\bfx)$.


Our work is based on the common implicit assumption of linear interpretability in the concept-based explanation literature, \ie high-level concepts lie in a linearly-projected subspace of the feature representation space $\calZ$ of the classifier~\citep{kim2018tcav}. 
Consider a projection matrix $\,\bfC = [\bfc_1, \cdots, \bfc_m] \in \reals^{d_\ell \times m}$ (with $m \ll d_\ell$) that maps from the space $\calZ$ into a reduced-dimension concept space. 
$\bfC$ consists of $m$ unit vectors, where $\bfc_i \in \reals^{d_\ell}$ is referred to as the \textit{concept vector} representing the $i$-th concept (\eg ``stripe'' or ``oval face''), and $m$ is the number of concepts.
We define the \textit{concept score} for $\bfx$ as the linear projection of the high-dimensional layer representation $\bfphi(\bfx) \in \reals^{a_\ell b_\ell \times d_\ell}$ into the concept space~\citep{yeh2020completeness}, i.e. $\,\VC(\bfx) := \bfphi(\bfx)\, \bfC \,\in \reals^{a_\ell b_\ell \times m}$.
We also define a mapping from the projected concept space back to the feature space by a non-linear function $\,\bfg : \reals^{a_\ell b_\ell \times m} \mapsto \reals^{a_\ell b_\ell \times d_\ell}$.
The reconstructed feature representation at layer $\ell$ is then defined as $\,\recphi(\bfx) \,:=\, \bfg(\VC(\bfx))$.

\subsection{Canonical World and Concept World}
\label{sec:two_worlds}
As shown in Fig.~\ref{fig:detection-completeness}, we consider a ``two-world'' view of the classifier and OOD detector consisting of the {\em canonical world} and the {\em concept world}, which are defined as follows:

\mypara{Canonical World.}
In this case, both the classifier and OOD detector use the original layer representation $\bfphi(\bfx)$ for their predictions. The prediction of the classifier is $\,\bff(\bfx) = \bfh(\bfphi(\bfx))$, and the decision function of the detector is $\,\calD_\gamma(\bfx, \bfh \circ \bfphi)$ with a score function $S(\bfx, \bfh \circ \bfphi)$.

\mypara{Concept World.}
We use the following observation in constructing the concept-world formulation: {\em both the classifier and the OOD detector can be modified to make predictions based on the reconstructed feature representation}, \ie using $\recphi(\bfx)$ instead of $\bfphi(\bfx)$. 
Accordingly, we define the corresponding classifier, detector, and score function in the concept world as follows:
\begin{align}
\label{equ:concept_world}
\centering
\fcon(\bfx) ~&:=~ \bfh(\recphi(\bfx)) ~=~ \bfh(\bfg(\VC(\bfx))) \nonumber \\
\Dcon(\bfx, \bff) \,&:=\, \calD_\gamma(\bfx, \bfh \circ \recphi) \,=\, \calD_\gamma(\bfx, \bfh \circ \bfg \circ \VC) \nonumber \\
\centering
\Scon(\bfx, \bff) \,&:=\, S(\bfx, \bfh \circ \recphi) \,=\, S(\bfx, \bfh \circ \bfg \circ \VC).
\end{align}
%
%
We further elaborate on this two-world view and introduce the following two desirable properties.



%
\mypara{Detection Completeness.} Given a fixed algorithmic approach for learning the classifier and OOD detector, and with fixed internal parameters of $\bff$, we would ideally like the classifier prediction and the detection score to be indistinguishable between the two worlds. 
In other words, for the concepts to \textit{sufficiently} explain the OOD detector, we require $\Dcon(\bfx, \bff)$ to closely mimic $\calD_\gamma(\bfx, \bff)$. 
Likewise, we require $\fcon(\bfx)$ to closely mimic $\bff(\bfx)$ since the detection mechanism of $\calD_\gamma$ is closely paired to the classifier.
We refer to this property as the {\em completeness of a set of concepts with respect to the OOD detector and its paired classifier.}  
As discussed in \S~\ref{sec:completeness_score}, this extends the notion of classification completeness introduced by~\citet{yeh2020completeness} to an OOD detector and its paired classifier.

\mypara{Concept Separability.} To improve the interpretability of the resulting explanations for the OOD detector, we require another desirable property from the learned concepts: data detected as ID by $\calD_\gamma$ (henceforth referred to as \textit{detected-ID} data) and data detected as OOD by $\calD_\gamma$ (henceforth referred to as \textit{detected-OOD} data) should be well-separated in the concept-score space.
Since our goal is to help an analyst understand which concepts distinguish the detected-ID data from detected-OOD data, we would like to learn a set of concepts that have a well-separated concept score pattern for inputs from these two groups (\eg the concepts $C_{90}$ and $C_1$ in Fig.~\ref{fig:expl-ours-dolphin} have distinct concept scores).

\vspace{-.05in}
\section{Proposed Approach}
Given a trained DNN classifier $\bff$, a paired OOD detector $\calD_\gamma$, and a set of concepts $\bfC$, we address the following questions: \textbf{1)} \emph{Are the concepts sufficient to capture the prediction behavior of both the classifier and OOD detector?} (see \S~\ref{sec:completeness_score}); \textbf{2)} \emph{Do the concepts show clear distinctions in their scores between detected-ID data and detected-OOD data?} (see \S~\ref{sec:separability_score}). 
We first propose new metrics for quantifying the set of learned concepts, followed by a general framework for learning concepts that possess these properties (see \S~\ref{sec:concept_learning}).

\subsection{Metric for Detection Completeness}
\label{sec:completeness_score}
%
%
%
\begin{definition}
\label{def:completeness_class}
Given a trained DNN classifier $\bff = \bfh \circ \bfphi\,$ and a set of concept vectors $\bfC$, the {\em classification completeness} with respect to $\Pin(\bfx, y)$ is defined as \citep{yeh2020completeness}:
\begin{align*}
    &\eta^{}_\bff(\bfC) := \\
    &\frac{\textrm{sup}_\bfg \,\expec_{(\bfx, y) \sim \Pin} \!\big[ \indicator[ y = \argmax_{y'} h_{y'}(\recphi(\bfx)) ] \big] ~-~ a_r}{\expec_{(\bfx, y) \sim \Pin} \!\big[ \indicator[y = \argmax_{y'} h_{y'}(\bfphi(\bfx))] \big] ~-~ a_r}
\end{align*}
where $a_r = 1 / L$ is the accuracy of a random classifier.
\end{definition}
The denominator of $\eta^{}_\bff(\bfC)$ is the accuracy of the original classifier $\bff$, while the numerator is the maximum accuracy that can be achieved by the concept-world classifier.
The maximization is over the parameters of the neural network $\bfg$ that reconstructs the feature representation from the vector of concept scores.

\begin{definition}
\label{def:completeness_detec}
Given a trained DNN classifier $\bff = \bfh \circ \bfphi$, a trained OOD detector with score function $S(\bfx, \bff)$, and a set of concept vectors $\bfC$, we define the {\em detection completeness score} with respect to the ID distribution $\Pin(\bfx, y)$ and OOD distribution $\Pout(\bfx)$ as follows:
\begin{align}
\label{equ: completeness-detection}
    \eta^{}_{\bff, S}(\bfC) 
    ~:=~ \frac{\textrm{sup}_\bfg \,\textrm{AUC}(\bfh \circ \recphi) ~-~ b_r}{\textrm{AUC}(\bfh \circ \bfphi) ~-~ b_r},
\end{align}
where $\textrm{AUC}(\bff)$ is the area under the ROC curve of an OOD detector based on $\bff$, defined as $\,\textrm{AUC}(\bff) \,:=\, \expec_{(\bfx, y) \sim \Pin} \expec_{\,\bfx^\prime \sim \Pout} \indicator\big[ S(\bfx, \bff) \,>\, S(\bfx^\prime, \bff) \big]$,
%
%
and $b_r = 0.5$ is the AUROC of a random detector.
\end{definition}

The numerator is the maximum achievable AUROC in the concept world using the reconstructed representation from concept scores.
In practice, $\textrm{AUC}(\bff)$ is estimated using the test datasets $\Dinte$ and $\Doutte$.
%
%
Both the classification completeness and detection completeness are designed to be in the range $[0, 1]$. However, this is not strictly guaranteed since the classifier or OOD detector in the concept world may empirically have a better (corresponding) metric on a given ID/OOD dataset.
Completeness scores close to $1$ indicate that the set of concepts $\bfC$ are close to complete in characterizing the behavior of the classifier and/or OOD detector.

\subsection{Concept Separability Score}
\label{sec:separability_score}
\mypara{Concept Scores.}
In Section \ref{sec:concept_projection}, we introduced a projection matrix $\bfC \in \reals^{d_\ell \times m}$ that maps $\bfphi(\bfx)$ to $\bfv_{\bfC}(\bfx)$, and consists of $m$ unit concept vectors $\,\bfC = [\bfc_1 \cdots \bfc_m]$. 
The inner product between the feature representation and a concept vector is referred to as the {\em concept score}, and it quantifies how close an input is to the given concept~\citep{kim2018tcav, ghorbani2019ace}.
%
Specifically, the concept score corresponding to concept $i$ is defined as $\,\bfv_{\bfc_i}(\bfx) := \langle \bfphi(\bfx), \bfc_i \rangle = \bfphi(\bfx) \,\bfc_i \in \reals^{a_\ell b_\ell}$. 
%
%
The matrix of concept scores from all the concepts is simply the concatenation of the individual concept scores, \ie $\VC(\bfx) = \bfphi(\bfx) \,\bfC = [\bfv_{\bfc_1}(\bfx) \cdots \bfv_{\bfc_m}(\bfx)] \in \reals^{a_\ell b_\ell \times m}$.
%
%
%
We also define a dimension-reduced version of the concept scores that takes the maximum of the inner-product over each $a_\ell \times b_\ell$ patch as follows: $\TVC(\bfx)^T = [\widetilde{v}_{\bfc_1}(\bfx), \cdots, \widetilde{v}_{\bfc_m}(\bfx)] \in \reals^m$, where $\,\widetilde{v}_{\bfc_i}(\bfx) = \max_{p, q} |\langle \bfphi^{p,q}(\bfx), \bfc_i \rangle| \in \reals$. Here $\bfphi^{p,q}(\bfx)$ is the feature representation corresponding to the $(p, q)$-th patch of input $\bfx$ (\ie receptive field~\citep{araujo2019computing}).
This reduction operation is done to capture the most important correlations from each patch, and the $m$-dimensional concept score will be used to define our concept separability metric as follows.
We would like the set of concept-score vectors from the detected-ID class $\,V_{\textrm{in}}(\bfC) := \{\TVC(\bfx), ~\bfx \in \Dintr \cup \Douttr ~:~ \calD_\gamma(\bfx, \bff) = 1\}$, and the set of concept-score vectors from the detected-OOD class $\,V_{\textrm{out}}(\bfC) := \{\TVC(\bfx), ~\bfx \in \Dintr \cup \Douttr ~:~ \calD_\gamma(\bfx, \bff) = 0\}\,$ to be well separated.
Let $\,J_{\textrm{sep}}(V_{\textrm{in}}(\bfC), V_{\textrm{out}}(\bfC)) \in \reals$ define a general {\it measure of separability} between the two subsets, such that a larger value corresponds to higher separability. We discuss a specific choice for $J_{\textrm{sep}}$ for which it is possible to tractably optimize concept separability as part of the learning objective in Section \ref{sec:concept_learning}.


\mypara{Global Concept Separability.} Class separability metrics have been well studied in the pattern recognition literature, particularly for the two-class case~\citep{fukunaga1990separ}\,\footnote{In our problem, the two classes correspond to ``detected-ID'' and ``detected-OOD''.}. 
%
Motivated by Fisher's linear discriminant analysis (LDA), we explore the use of class-separability measures based on the within-class and between-class scatter matrices~\citep{murphy2012separ}.
The goal of LDA is to find a projection vector (direction) such that data from the two classes are maximally separated and form compact clusters upon projection. 
Rather than finding an optimal projection direction, we are more interested in ensuring that the concept-score vectors from the detected-ID and detected-OOD data have high separability.
Consider the within-class and between-class scatter matrices based on $V_{\textrm{in}}(\bfC)$ and $V_{\textrm{out}}(\bfC)$, given by
\begin{align}
\label{eq:scatter_matrices}
\bfS_w ~&= \mysum_{\bfv \in V_{\textrm{in}}(\bfC)} (\bfv \,-\, \bfmu_{\textrm{in}})\,(\bfv \,-\, \bfmu_{\textrm{in}})^T \nonumber \\
&\,+\, \mysum_{\bfv \in V_{\textrm{out}}(\bfC)} (\bfv \,-\, \bfmu_{\textrm{out}})\,(\bfv \,-\, \bfmu_{\textrm{out}})^T, \\
\bfS_b ~&=~ (\bfmu_{\textrm{out}} \,-\, \bfmu_{\textrm{in}})\,(\bfmu_{\textrm{out}} \,-\, \bfmu_{\textrm{in}})^T,
\end{align}
where $\bfmu_{\textrm{in}}$ and $\bfmu_{\textrm{out}}$ are the mean concept-score vectors from $V_{\textrm{in}}(\bfC)$ and $V_{\textrm{out}}(\bfC)$ respectively.
We define the following separability metric based on the generalized eigenvalue equation solved by Fisher's LDA~\citep{fukunaga1990separ}: $J_{\textrm{sep}}(\bfC) ~:=~ J_{\textrm{sep}}(V_{\textrm{in}}(\bfC), V_{\textrm{out}}(\bfC)) ~=~ \textrm{tr}\big[\bfS_w^{-1} \,\bfS_b\big]$.
%
%
Maximizing the above metric is equivalent to maximizing the sum of eigenvalues of the matrix $\,\bfS_w^{-1} \,\bfS_b$, which in-turn ensures a large between-class separability and a small within-class separability for the detected-ID and detected-OOD concept scores.
We refer to this as a {\em global concept separability} metric because it does not analyze the separability on a per-class level.
The separability metric is closely related to the Bhattacharya distance, which is an upper bound on the Bayes error rate (see Appendix \ref{sec:appendix-BC-distance}).
We define the per-class variations of detection completeness and concept separability in a similar way in Appendix \ref{sec:appendix-perclass-completeness} and \ref{sec:appendix-perclass-separability}.

\subsection{Proposed Concept Learning -- Key Ideas}
\label{sec:concept_learning}

\mypara{Prior Approaches and Limitations.} 
Among post-hoc concept-discovery methods for a DNN classifier with ID data, 
unlike \citeauthor{kim2018tcav} and \citeauthor{ghorbani2019ace}, that do not support imposing required conditions into the concept discovery, \citeauthor{yeh2020completeness} devised a learning-based approach where classification completeness and the saliency of concepts are optimized via a regularized objective given by
\vspace{-1mm}
\begin{equation}
\label{equ: baseline}
    \argmax_{\bfC, \bfg} \expec_{(\bfx, y) \sim \Pin}\!\!\big[ \log h_y(\bfg(\VC(\bfx))) \big] ~+~ \lambda_{\textrm{expl}}\, R_{\textrm{expl}}(\bfC).
\end{equation}
Here $\bfC$ and $\bfg$ (parameterized by a neural network) are jointly optimized, and $R_{\textrm{expl}}(\bfC)$ is a regularization term used to ensure that the learned concept vectors have high spatial coherency and low redundancy among themselves (see \citet{yeh2020completeness} for the definition).
%

\begin{figure*}[t]
  \centering
  \begin{subfigure}{0.32\linewidth}
    \includegraphics[width=\textwidth]{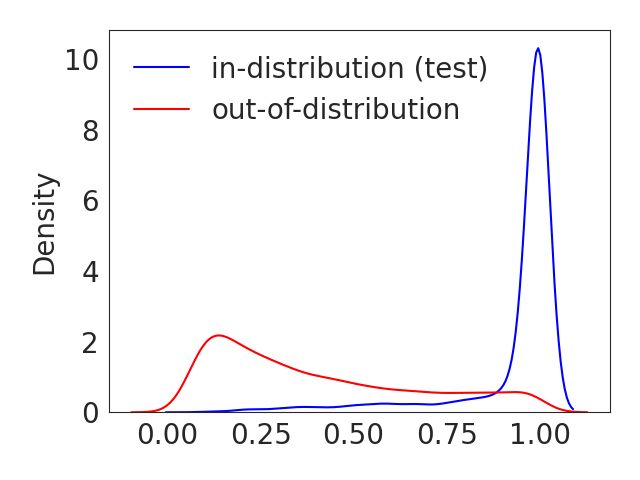}
    \caption{\small Empirical distribution of $S(\bfx, \bff)$ from the target detector.}
    \label{fig:short-a}
  \end{subfigure}
  \hfill
  \begin{subfigure}{0.32\linewidth}
    \includegraphics[width=\textwidth]{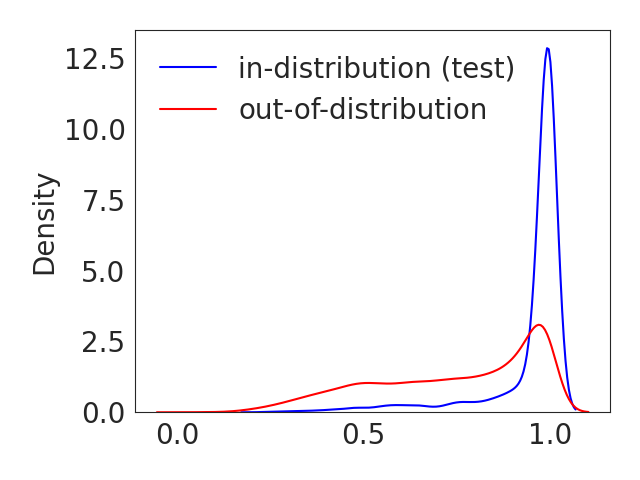}
    \caption{\small Distribution of $\Scon(\bfx, \bff)$ using the concepts learned by \citet{yeh2020completeness}.}
    \label{fig:short-b}
  \end{subfigure}
  \hfill
  \begin{subfigure}{0.32\linewidth}
    \includegraphics[width=\textwidth]{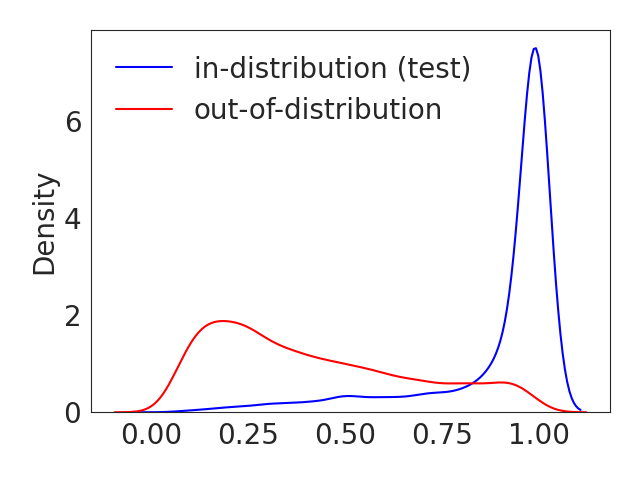}
    \caption{\small Distribution of $\Scon(\bfx, \bff)$ using the concepts learned by our method.}
    \label{fig:short-c}
  \end{subfigure}
  \caption{\small Empirical distribution of: \textbf{(a)} MSP detector score $S(\bfx, \bff)$ in the canonical world vs. \textbf{(b, c)} Reconstructed $\Scon(\bfx, \bff)$ in the concept world using the concepts learned by \citet{yeh2020completeness} and our method.
  Concepts learned by~\citet{yeh2020completeness} have $\eta^{}_{\bff} = 0.977, ~\eta^{}_{\bff, S}(\bfC) = 0.782$, while the concepts learned by our method $(\lambda_\textrm{mse} = 10, \lambda_\textrm{norm} = 0.1, \lambda_\textrm{sep} = 50)$ have $\eta^{}_{\bff} = 0.984, ~\eta^{}_{\bff, S}(\bfC) = 0.961$. 
  The AwA test set and the \texttt{SUN} dataset are used as ID (blue) and OOD (red) respectively.
    }
    \vspace{-.1in}
\label{fig:score-distribution-msp}
\end{figure*}

While the objective (\ref{equ: baseline}) of \citeauthor{yeh2020completeness} can learn a set of sufficient concepts that have a high classification completeness score, we find that it does not necessarily replicate the per-instance prediction behavior of the classifier in the concept world. 
Specifically, there can be discrepancies in the reconstructed feature representation, whose effect propagates through the remaining part of the classifier. 
Since many widely-used OOD detectors rely on the feature representations and/or the classifier's predictions, this discrepancy in the existing concept learning approaches makes it hard to closely replicate the OOD detector in the concept world (see Fig.~\ref{fig:score-distribution-msp}).
Furthermore, the scope of \citeauthor{yeh2020completeness} is confined to concept learning for explaining the classifier's predictions based on ID data, and there is no guarantee that the learned concepts would be useful for explaining an OOD detector. 
%
To address these gaps, we propose a general method for concept learning that complements prior work by imposing additional instance-level constraints on the concepts, and by considering both the OOD detector and OOD data.

\mypara{Concept Learning Objective.}
We define a concept learning objective that aims to find a set of concepts $\bfC$ and a mapping $\bfg$ that have the following properties: 1) high detection completeness w.r.t the OOD detector; 2) high classification completeness w.r.t the DNN classifier; and 3) high separability in the concept-score space between detected-ID data and detected-OOD data.

Inspired by recent works on transferring feature information from a teacher model to a student model~\citep{hinton2015distilling,zhou2018rocket},
we encourage accurate reconstruction of $\hat{\calZ}$ based on the concept scores by adding a regularization term that is the squared $\ell_2$ distance between the original and reconstructed representations $\,J_{\textrm{norm}}(\bfC, \bfg) ~=~ \expec_{\bfx \sim \Pin} \|\bfphi(\bfx) \,-\, \recphi(\bfx)\|^2\,$.
%
%
In order to close the gap between the scores of the OOD detector in the concept world and canonical world on a per-sample level, we introduce the following mean-squared-error (MSE) based regularization:
\begin{align}
\label{equ:regularizer_mse}
    J_{\textrm{mse}}(\bfC, \bfg) ~&=~ \expec_{\bfx \sim \Pin} \big( S(\bfx, \bfh \circ \recphi) - S(\bfx, \bff) \big)^2 \nonumber \\
    &~+~ \expec_{\bfx \sim \Pout} \big( S(\bfx, \bfh \circ \recphi) - S(\bfx, \bff) \big)^2 .
\end{align}
MSE terms are computed with both the ID and OOD data because we want to ensure that the ROC curve corresponding to both the score functions are close to each other (which requires OOD data).
Finally, we include a regularization term to maximize the separability metric between the detected-ID and detected-OOD data in the concept-score space, resulting in our final concept learning objective:
\begin{align}
\label{equ: concept learning}
&\argmax_{\bfC, \bfg}  \expec_{(\bfx, y) \sim \Pin}\!\!\big[ \log h_y(\bfg(\VC(\bfx))) \big] ~+~ \lambda_{\textrm{expl}}\, R_{\textrm{expl}}(\bfC) \nonumber \\ 
    &-~ \lambda_{\textrm{mse}}\, J_{\textrm{mse}}(\bfC, \bfg) ~-~ \lambda_{\textrm{norm}}\, J_{\textrm{norm}}(\bfC, \bfg) ~+~ \lambda_{\textrm{sep}}~J_{\textrm{sep}}(\bfC).
\end{align}
The $\lambda$ coefficients are non-negative hyper-parameters that are further discussed in Section~\ref{subsec:results}.
We note that both $J_{\textrm{mse}}(\bfC, \bfg)$ and $J_{\textrm{sep}}(\bfC)$ depend on the OOD detector~\footnote{This dependence may not be obvious for the separability term, but it is clear from its definition.}.
We use the SGD-based Adam optimizer~\citep{kingma2014adam}) to solve the learning objective.
The expectations involved in the objective terms are calculated using sample estimates from the training ID and OOD datasets. 
Specifically, $\Dintr$ and $\Douttr$ are used to compute the expectations over $\Pin$ and $\Pout$, respectively.
Our complete concept learning is summarized in Algorithm \ref{alg:concept_learning} (Appendix \ref{sec:appendix-algo}).

\section{Experiments}
\label{subsec:results}

In this section, we conduct experiments to evaluate the proposed method and show that: 1) the learned concepts satisfy the desiderata of completeness and separability across popular off-the-shelf OOD detectors and real-world datasets. 2) the learned concepts can be combined with a Shapley value to provide insightful visual explanations that can help understand the predictions of an OOD detector. 
The code for our work can be found at \url{https://github.com/jihyechoi77/concepts-for-ood}.

\subsection{Experimental Setup}
\label{sec:setup}

\mypara{Datasets.} For the ID dataset, we use Animals with Attributes (AwA) \citep{xian2018awa} with 50 animal classes, and split it into a train set (29841 images), validation set (3709 images), and test set (3772 images).
We use the MSCOCO dataset \citep{lin2014mscoco} as the auxiliary OOD dataset $\Douttr$ for training and validation.
For the OOD test dataset $\Doutte$, we follow the literature of large-scale OOD detection \citep{Huang_MOS} and use three different image datasets: \texttt{Places365} \citep{zhou2017places}, \texttt{SUN} \citep{xiao2010sun}, and \texttt{Textures} \citep{cimpoi2014textures}.



\mypara{Models.}
We apply our framework to interpret five prominent OOD detectors from the literature: MSP~\citep{hendrycks2016msp}, ODIN~\citep{liang2018ODIN}, Generalized-ODIN~\citep{hsu2020GeneralizedODIN}, Energy~\citep{liu2020energy}, and Mahalanobis~\citep{lee2018mahalanobis}.
The OOD detectors are paired with the widely-used Inception-V3 classifier~\citep{szegedy2016inception-v3} (following the setup in prior works~\citep{yeh2020completeness, ghorbani2019ace, kim2018tcav}), which has a test accuracy of $92.13 \%$ on the AwA dataset.
%
Additional details on the setup are given in Appendix~\ref{sec:appendix-implementation-details}.
 
\subsection{Effectiveness of the Concept Learning}
\label{sec:eval-concept}
Table~\ref{tab:concept-learning-results-places} summarizes the results of concept learning for various combinations of the regularization coefficients ($\lambda_{\star}$) in Eqn (\ref{equ: concept learning}), including: \textbf{i)} baseline where all the coefficients are set to $0$ (first row); \textbf{ii)} only the terms directly relevant to detection completeness (\ie $J_{\textrm{norm}}(\bfC, \bfg)$ and $J_{\textrm{mse}}(\bfC, \bfg)$) are included (second row); \textbf{iii)} only the term responsible for concept separability $J_{\textrm{sep}}(\bfC)$ is included (third row); \textbf{iv)} all the regularization terms are included (fourth row).

From the table, we observe that the regularization terms encourage the learned concepts to satisfy the required desiderata of high completeness and concept separability scores.
Having $\lambda_\textrm{mse} > 0\, \text{ and } \,\lambda_\textrm{norm} > 0$ always improves the detection completeness by a large margin (\ie row 2 compared to row 1), and having $\lambda_\textrm{sep} > 0$ significantly increases the concept separability (\ie row 3 compared to row 1).
Importantly, when all the regularization terms are included, they have the best synergistic effect on the metrics. 

\begin{table}[thb]
    \centering
    \begin{adjustbox}{width=1\columnwidth,center}
		\begin{tabular}{l|l|c|c|c}
			\toprule
			OOD detector & $(\lambda_\textrm{mse}, \lambda_\textrm{norm}, \lambda_\textrm{sep})$ &
			$\eta^{}_{\bff}(\bfC) \uparrow$ & $\eta^{}_{\bff, S}(\bfC) \uparrow$ & $J_{\textrm{sep}}(\bfC, \bfC') \uparrow$ \\ \hline \hline
            \multirow{4}{0.10\linewidth}{MSP} 
			& $(0, 0, 0)$ & 0.977 $\pm$ 0.0006 & 0.774 $\pm$ 0.0010 & 0.694 $\pm$ 0.0153 \\
			& $(10, 0.1, 0)$ & \textbf{0.994} $\pm$ 0.0004 & \underline{0.947} $\pm$ 0.0004 & 1.892 $\pm$ 0.0393 \\
			& $(0, 0, 50)$ & 0.980 $\pm$ 0.0005 & 0.814 $\pm$ 0.0008 & \underline{2.533} $\pm$ 0.0714 \\
			& $(10, 0.1, 50)$ & \underline{0.984} $\pm$ 0.0004 & \textbf{0.960} $\pm$ 0.0004 & \textbf{2.756} $\pm$ 0.0854 \\ \hline
            \multirow{4}{0.10\linewidth}{ODIN} 
			& $(0, 0, 0)$ & 0.977 $\pm$ 0.0006 & 0.742 $\pm$ 0.0011 & 0.444 $\pm$ 0.0119 \\
			& $(10^8, 0.1, 0)$ & \textbf{0.994} $\pm$ 0.0004 & \underline{0.951} $\pm$ 0.0004 & 1.166 $\pm$ 0.0303 \\
			& $(0, 0, 50)$ & 0.987 $\pm$ 0.0004 & 0.899 $\pm$ 0.0007 & \underline{1.785} $\pm$ 0.0669 \\
			& $(10^8, 0.1, 50)$ & \underline{0.991} $\pm$ 0,0005 & \textbf{0.973} $\pm$ 0.0009 & \textbf{1.813} $\pm$ 0.0268 \\ \hline
            \multirow{4}{0.10\linewidth}{General-ODIN} 
			& $(0, 0, 0)$ & 0.988 $\pm$ 0.0004 & 0.769 $\pm$ 0.0004 & 0.506 $\pm$ 0.0165 \\
			& $(10^6, 0.1, 0)$ & \textbf{0.995} $\pm$ 0.0004 & \underline{0.951} $\pm$ 0.0006 & 1.461 $\pm$ 0.0321 \\
			& $(0, 0, 50)$ & 0.981 $\pm$ 0.0004 & 0.859 $\pm$ 0.0007 & \underline{1.814} $\pm$ 0.0685 \\
			& $(10^6, 0.1, 50)$ & \underline{0.990} $\pm$ 0.0005 & \textbf{0.971} $\pm$ 0.0010 & \textbf{1.835} $\pm$ 0.0669 \\ \hline
            \multirow{4}{0.10\linewidth}{Energy} 
			& $(0, 0, 0)$ & 0.977 $\pm$ 0.0006 & 0.671 $\pm$ 0.0012 & 0.453 $\pm$ 0.0121 \\
			& $(1. 0.1, 0)$ & \textbf{0.993} $\pm$ 0.0005 & \textbf{0.965} $\pm$ 0.0004 & 1.266 $\pm$ 0.0319 \\
			& $(0, 0, 50)$ & \underline{0.987} $\pm$ 0.0005 & 0.779 $\pm$ 0.0010 & \textbf{1.920} $\pm$ 0.0725 \\
			& $(1, 0.1, 50)$ & 0.980 $\pm$ 0.0005 & \underline{0.943} $\pm$ 0.0005 & \underline{1.839} $\pm$ 0.0662 \\ \hline
			\multirow{4}{0.10\linewidth}{Mahalanobis} 
			& $(0, 0, 0)$ & 0.990 $\pm$ 0.0007 & 0.715 $\pm$ 0.0011 & 0.571 $\pm$ 0.0110 \\
			& $(0.1, 0.1, 0)$ & \textbf{0.994} $\pm$ 0.0004 & \underline{0.950} $\pm$ 0.0009 & 1.532 $\pm$ 0.0351 \\
			& $(0, 0, 50)$ & 0.985 $\pm$ 0.0004 & 0.880 $\pm$ 0.0005 & \underline{2.550} $\pm$ 0.0681 \\
			& $(0.1, 0.1, 50)$ & \underline{0.992} $\pm$ 0.0006 & \textbf{0.961} $\pm$ 0.0005 & \textbf{2.616} $\pm$ 0.0857 \\ \bottomrule
		\end{tabular}
	\end{adjustbox}
	\caption[]{
	\small Concept learning results with different parameter settings across various OOD detectors, evaluated on AwA test set (ID) and \texttt{Places365} (OOD). 
Hyperparameters are set based on the scale of corresponding regularization terms for a specific choice of the OOD detector.
 Across the rows (for a given OOD detector), the best value is \textbf{boldfaced}, and the second best value is \underline{underlined}.
	The $95\%$ confidence intervals are estimated by bootstrapping the test set over $200$ trials. Complete results are given in Table~\ref{tab:concept-learning-results} in Appendix~\ref{sec:app_addi_results}.
	}
\label{tab:concept-learning-results-places}
\end{table}
%

Consider the MSP detector for instance. The detection completeness increases from $0.774$ to $0.947$ with $\lambda_\textrm{mse} = 10, \lambda_\textrm{norm} = 0.1, \lambda_\textrm{sep} = 0$, and the concept separability increases from $0.694$ to $2.533$ with $\lambda_\textrm{mse} = 0, \lambda_\textrm{norm} = 0, \lambda_\textrm{sep} = 50$.
However, when all the terms are considered ($\lambda_\textrm{mse} = 10, \lambda_\textrm{norm} = 0.1, \lambda_\textrm{sep} = 50$), we achieve the best result of $\eta^{}_{\bff, S}(\bfC) = 0.960$ and $J_{\textrm{sep}}(\bfC, \bfC') = 2.756$. 
Results on other large-scale OOD datasets and ablation studies on the regularization terms can be found in Appendix \ref{sec:appendix-concept-learning-ablation}.

Since the range of the separability score $J_{\textrm{sep}}(\bfC)$ (or $J^y_{\textrm{sep}}(\bfC)$) is not well defined, we report a \textit{relative concept separability} score that is easier to interpret, and defined as
\begin{equation}
\label{eq:relative-separability}
J_{\textrm{sep}}(\bfC, \bfC') = \textrm{Median}\left( \left\{\frac{J^y_{\textrm{sep}}(\bfC) ~-~ J^y_{\textrm{sep}}(\bfC')}{J^y_{\textrm{sep}}(\bfC')}\right\}_{y=1}^L \right). 
\end{equation}
It captures the relative improvement in concept separability using concepts $\bfC$, compared to a baseline set of concepts $\bfC'$ obtained by setting $\,\lambda_\textrm{mse} = \lambda_\textrm{norm} = \lambda_\textrm{sep} = 0$.
%

\mypara{Concepts good for the OOD detector are also good for the classifier, but not vice-versa.} 
Recall that the baseline $(\lambda_\textrm{mse} = \lambda_\textrm{norm} = \lambda_\textrm{sep} = 0)$ corresponds to the method of \citet{yeh2020completeness} where only the classifier is considered during the concept learning.
For any choice of OOD detector in Table~\ref{tab:concept-learning-results-places} and Table~\ref{tab:concept-learning-results} (in Appendix~\ref{sec:app_addi_results}), the concepts learned by our method always achieve higher scores even for classification completeness, compared to the baseline. 
In contrast, the baseline concepts for only the classifier have the lowest detection completeness and concept separability in all cases.
This may not be surprising since the scope of~\citet{yeh2020completeness} does not cover explaining detectors.
Nonetheless, such observations provide supporting evidence to motivate the need for our concept learning and novel metrics, indicating that {\em even if the concepts are sufficient to describe the DNN classifier, the same set of concepts may not be appropriate for the OOD detector.}

\begin{figure*}[hbt]
     
     \begin{subfigure}[b]{0.7\textwidth}
     \centering
     \begin{subfigure}{\textwidth}
         \centering
         \includegraphics[width=\textwidth]{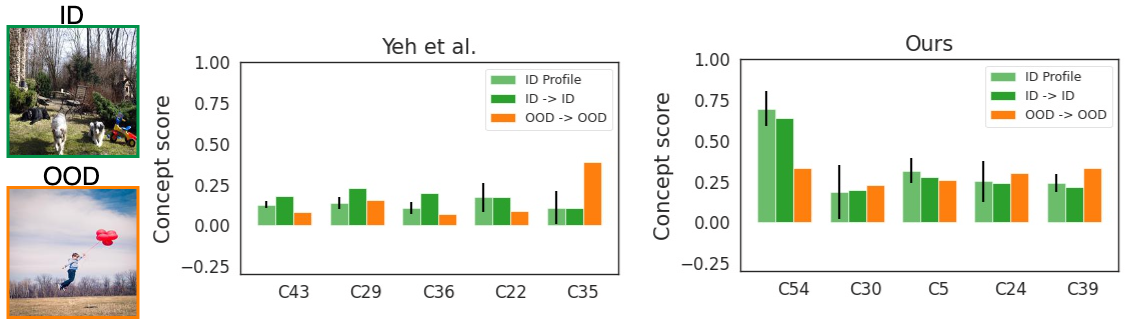}
         \caption{\small Correct detection: top collie image is correctly detected as ID (dark-green bar), and the bottom image is correctly detected as OOD (orange bar).}
         \label{fig:collie-correct}
     \end{subfigure}
     \\
     \begin{subfigure}{\textwidth}
         \centering
         \includegraphics[width=\textwidth]{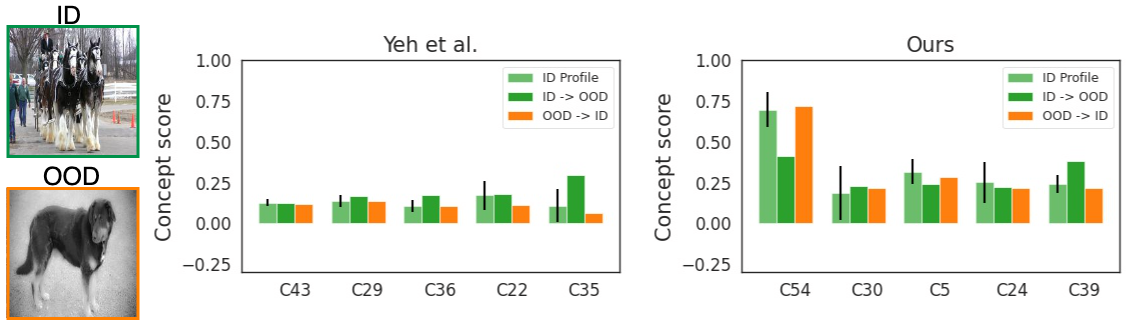}
         \caption{\small Wrong detection: top ID image is detected as OOD (dark-green bar), and the bottom OOD image is detected as ID (orange bar).}
         \label{fig:collie-wrong}
     \end{subfigure}
     \end{subfigure}
     \hspace{2mm}
     \begin{subfigure}[b]{0.23\textwidth}
         \centering
         \includegraphics[width=\textwidth]{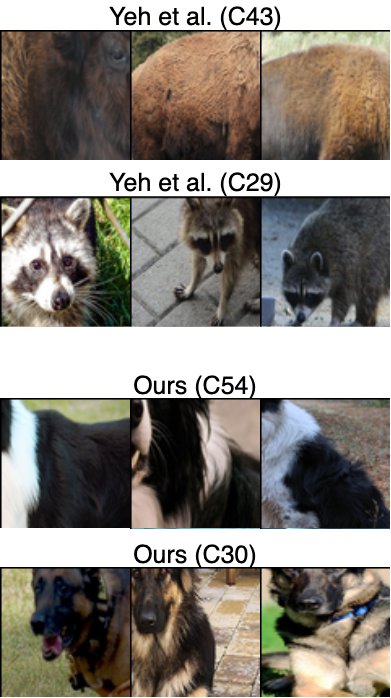}
         \caption{\small Visualization of top-2 important concepts found by the method of \citet{yeh2020completeness} and our method.}
         \label{fig:collie-concepts}
     \end{subfigure}
\caption{\small \textbf{Concept-based explanations for the Energy OOD detector using concepts learned by \citet{yeh2020completeness} vs. ours ($\lambda_\textrm{mse} = 1, \lambda_\textrm{norm} = 0.1, \lambda_\textrm{sep} = 10$)}. Images are randomly selected from the AwA test set (ID) and \texttt{Places} (OOD), and all of them are predicted to the class ``Collie''. Concept score patterns by \citet{yeh2020completeness} are not distinctive between detected-ID vs. detected-OOD (\ie dark-green bar and orange bar are not very different). Whereas, our concepts present very similar patterns to the ID profile (light-green bar) when input is detected as ID, and different from the ID profile when detected as OOD.
}
\label{fig:expl-energy-collie}

\end{figure*}


\mypara{Accurate Reconstruction of Per-sample Behavior.}
In addition to the above numerical comparisons with respect to the proposed metrics, we found the method of \citet{yeh2020completeness} to have potential issues in terms of reconstructing the feature representations. This in-turn leads to degraded reconstruction of the \textit{per-sample} behavior of the OOD detector.
Comparing Fig.~\ref{fig:short-a} and Fig.~\ref{fig:short-b}, we observe that the concepts of \citet{yeh2020completeness} lead to a strong mismatch between the score distributions of the OOD detector. In contrast, our method approximates the score distributions more closely (compare Fig.~\ref{fig:short-a} and Fig.~\ref{fig:short-c}).
Given that the second half of the classifier and detector remains fixed between the canonical and concept worlds, this observation implies that the reconstructed features fed into the second half of the classifier have to be distorted. Similar observations are made for the Energy detector in Fig.~\ref{fig:score-distribution-energy} in Appendix~\ref{sec:app_addi_results}.
We observe that such inaccurate reconstruction of features poses a similar problem for classifiers as well (more discussion in Appendix~\ref{app:hellinger}).
We conclude that the objective of \citet{yeh2020completeness}, which considers only the aggregate statistic of reconstructed accuracy, is not sufficient to recover the per-sample behavior, and augmenting it with our reconstruction error-based regularization term is a straightforward improvement for both the classifier and OOD detector.

\subsection{Concept-Based Explanations for OOD Detectors}
\label{sec:expt_concept_based_explanations}

\mypara{Finding the Key Concepts.}
Given a set of learned concepts, we address the question: {\em how much does each concept contribute to the detection results for inputs predicted to a particular class?}
To address this, we follow recent works that have adopted the Shapley value from Game theory literature~\citep{shapley1953value,fujimoto2006axiomatic} for scoring the importance of a feature subset towards the predictions of a model~\citep{chen2018shapley,lundberg2017shapley,sundararajan2020shapley}.
We propose to use our per-class detection completeness metric $\eta^{j}_{\bff, S}(\bfC)\,$ (Eqn. (\ref{equ: completeness-detection-perclass}) in Appendix \ref{sec:appendix-perclass-completeness}) as the characteristic function of the Shapley value. 
The modified Shapley value of a concept $\bfc_i \in \bfC$ with respect to the predicted class $j \in [L]$ is defined as
\begin{align}
    \label{equ: ConceptSHAP}
    \textrm{SHAP}(\eta^{j}_{\bff, S}, \bfc_i) ~:= \!\!\!\sum_{\bfC' \subseteq \bfC \setminus \{\bfc_i\}} \!\!\! \frac{ \eta^{j}_{\bff, S}\big( \bfC' \cup \{\bfc_i\} \big) ~-~ \eta^{j}_{\bff, S}(\bfC') }{ m\, \binom{m \,-\, 1}{|\bfC'|} },
\end{align}
where $\bfC'$ is a subset of $\bfC$ excluding concept $\bfc_i$.
This Shapley importance score captures the average marginal contribution of concept $\bfc_i$ towards explaining the decisions of the OOD detector for inputs predicted into class $j$.

In the rest of the section, we demonstrate how the concepts ranked by the above Shapley importance score can serve as a useful tool for interpreting the OOD detector.

\mypara{Explaining Detection Errors.}
Given an OOD detector of interest, we collect inputs that are correctly detected as ID, and average their concept scores (which corresponds to the \textit{ID profile} in Fig.~\ref{fig:expl-energy-collie}).
The ID profile quantifies how much each concept matters for the normal ID inputs.
Given a test input, either correctly or incorrectly detected, the user could examine how similar or different this input is with respect to the ID concept profile.
Fig.~\ref{fig:expl-energy-collie} illustrates our explanations of the Energy OOD detector's decisions.
By visualizing the concepts (see Fig.~\ref{fig:collie-concepts}), we observe that for the predicted class \textit{Collie}, ``furry dog skin'' (C54) and ``oval dog face'' (C30) are the key concepts to capture the detector's outputs to distinguish ID images from OOD images.
We also observe that the OOD detector predicts an input as ID when the concept scores show a similar pattern to the ID profile, or predicts an input as OOD when the concept-score pattern is far from the ID profile.
For instance, our analysis shows that the bottom input in Fig.~\ref{fig:collie-wrong} is an OOD image from \texttt{Places} dataset but detected as ID (false positive) since its score for ``furry dog skin'' is as high as the usual ID Collie images (which is true in the image). 
Explaining detection results is crucial for encouraging the adoption of OOD detectors in various decision-making processes.
Our example here suggests that certain errors of an OOD detector can be understandable mistakes, which require further reasoning, rather than discarding the model based only on aggregate performance  metrics.
Additional examples of our concept-based explanations are given in Appendix~\ref{app:more-expl}.

\mypara{Comparison of Explanations by~\citeauthor{yeh2020completeness} and Ours.}
Lastly, we provide qualitative evidence supporting our argument that: concepts good for the classifier are not necessarily good for the OOD detector.
In Fig.~\ref{fig:expl-energy-collie}, given an Energy detector and ID/OOD inputs, we present explanations using  concepts learned by \citet{yeh2020completeness} vs. our method.
We observe that~\citet{yeh2020completeness} fails to generate visually-distinguishable explanations between detected-ID and detected-OOD inputs.
The separation between the dark-green bars and the orange bars in Fig.~\ref{fig:collie-correct} and Fig.~\ref{fig:collie-wrong} becomes more visible in our explanations, which enables more intuitive interpretation for human users (this reflects our design goal of concept separability).
It is also noteworthy that in Fig.~\ref{fig:collie-concepts}, our concepts that are most important to distinguish ID Collie from OOD Collie (\ie C54 and C30) are more specific and finer-grained characteristics of Collie, while \citet{yeh2020completeness} finds concepts that are vaguely similar to the features of a dog, but rather generic (\ie C43 and C29).
This is the reason we require more number of concepts to achieve high detection completeness and concept separability, compared to solely considering the classification completeness~\footnote{In Fig.~\ref{fig:expl-energy-collie}, after concept learning with $m = 100$ and duplicate removal, we found 44 non-redundant concepts for ~\citet{yeh2020completeness}, and 100 distinct concepts for ours.}.

\subsection{Explanations For Better OOD Detection.}
Our work makes the first effort to reason about the different failure modes of OOD detectors through explanations, rather than just observing aggregated performance metrics (e.g., AUROC or AUPRC).
Naturally, the next step would be to utilize such reasoning to modify and improve the OOD detector. 
We leave the development of concept-based explanations as actionable guidelines for better OOD detection as future work, and describe here a scenario where our explanations can provide direct utility.

We posit that our explanations can provide effective feedback when the failure of the OOD detector originates from misbehavior of the paired classifier, which we confirm to be the most common failure mode of OOD detectors. 
For instance, consider the top image in Fig.~\ref{fig:collie-wrong} as our input. Its true label is ``Horse'', but the classifier predicted it to class ``Collie''. Obviously, the horse image has a different concept-activation pattern from the normal ID Collie profile (compare the dark-green bars with the ID profile of ours in Fig.~\ref{fig:collie-wrong}). 
To remove such failure cases, the practitioner could identify the key concepts for the prediction of class ``Horse'' and compare the concept pattern of the input to the normal ID ``Horse'' profile. It is noteworthy that we can use the same set of concepts here, since our concept-learning objective finds concepts that can effectively explain \textit{both} the classifier and OOD detector. Indeed, we observe that the key concepts for the class ``Horse'' are ``round brown body'' and ``brown oval face of horse'', while the given input is an outlier relative to these concepts. 
Hence, the practitioner could consider diversifying the training set in the ID ``Horse'' class to include more examples of horses (\eg with black and white hair).

\section{Conclusion}
We develop an unsupervised and human-interpretable explanation method for black-box OOD detectors based on high-level concepts derived from the internal layer representations of a (paired) DNN classifier.
We propose novel metrics viz. {\em detection completeness} and {\em concept separability} to evaluate the completeness (sufficiency) and quality of the learned concepts for OOD detection.
We then propose a concept learning method that is quite general and applies to a broad class of off-the-shelf OOD detectors.
Through extensive experiments and qualitative examples, we demonstrate the practical utility of our method for understanding and debugging OOD detectors.
We discuss additional aspects of our method such as the auxiliary OOD dataset, human subject study, and societal impact in Appendix~\ref{sec:app_discussion}.

\section*{Acknowledgements}
We thank all the anonymous reviewers for their careful comments and feedback.
The work is partially supported by Air Force Grant FA9550-18-1-0166, the National Science Foundation (NSF) Grants CCF-FMitF-1836978, IIS-2008559, SaTC-Frontiers-1804648, CCF-2046710, CCF-1652140, and 2039445, and ARO grant number W911NF-17-1-0405. Choi, Feng, Chen, Jha, and Prakash are partially supported by the DARPA-GARD problem under agreement number 885000.
Raghuram is partially supported through the NSF grants CNS-2112562, CNS-2107060, CNS-2003129, CNS-1838733, CNS-1647152, and the US Department of Commerce grant 70NANB21H043.

\bibliographystyle{icml2023}
\bibliography{references}

\newpage
\appendix
\onecolumn
\begin{center}
	\textbf{\LARGE Appendix }
\end{center}

In Section~\ref{sec:app_discussion}, we discuss additional aspects of our method such as the choice of auxiliary OOD dataset, human subject study, and societal impact.
In Section \ref{sec:appendix-A}, we discuss the connection of the proposed concept separability to Bhattacharya Distance, and the per-class variations of detection completeness and concept separability, followed by the overall algorithm for concept learning.
In Section \ref{sec:appendix-implementation-details}, we provide the detailed setup for the experiments and additional thorough analysis of our concept learning objective.
In Section \ref{app:auxiliary-ood}, we discuss whether our concept learning objective remains effective even when a synthesized auxiliary OOD dataset similar to target ID data is used.
In Section \ref{sec:appendix-shapleys}, we illustrate additional examples of our concept-based explanations.

\section{Discussion and Societal Impact}
\label{sec:app_discussion}
\mypara{Auxiliary OOD Dataset.} 
A limitation of our approach is its requirement of an auxiliary OOD dataset for concept learning, which could be hard to access in certain applications.
To overcome that, a research direction would be to design generative models that simulate domain shifts or anomalous behavior and could create the auxiliary OOD dataset synthetically, allowing us additional control on the extent of distributional changes the resulting concepts could deal with (see Appendix~\ref{app:auxiliary-ood} for further discussion). 


\mypara{Human Subject Study.} 
Performing a human-subject (or user) study would be the ultimate way to evaluate the effectiveness of explanations, but remains largely unexplored even for in-distribution classification tasks.  
We emphasize that designing such a usability test with OOD detectors would be even more challenging due to the characteristics of the OOD detection task, compared to in-indistribution classification tasks.
For in-distribution classifiers, users could potentially generate hypotheses about what high-level concepts should attribute to the class prediction, and compare their hypotheses to the provided explanations to determine the classifier's reliability.
On the other hand, assessing the reliability of OOD detection involves checking whether a given input belongs to any of the natural distributions of concepts; this is essentially limited to whether users' mental models on such global distributions can be accurately probed via a couple of presented local instances.
We believe that designing a thorough probing method for human interpretability on OOD detection would be an interesting yet challenging research quest by itself~\cite{kim2022hive} and our paper does not address that.

\mypara{Societal Impact.}
\label{sec:broader_impact}
Our work helps address the detection results of OOD detectors, giving practitioners the ability to explain the model's decision to invested parties. 
Our explanations can also be used to keep a data point as an understood mistake by the model rather than throwing it away without further analysis, which could help guide how to improve the OOD detector with respect to the concepts. 
However, this would also mean that more trust is put back into the human practitioner to not abuse the explanations or misrepresent them. 

\section{Concept Learning}
\label{sec:appendix-A}
\subsection{Connection to the Bhattacharya Distance}
\label{sec:appendix-BC-distance}
We note that the proposed separability metric in Section \ref{sec:separability_score} is closely related to the Bhattacharya distance~\cite{bhattacharyya1943measure} for the special case when the concept scores from both ID and OOD data follow a multivariate Gaussian density. The Bhattacharya distance is a well known measure of divergence between two probability distributions, and it has the nice property of being an upper bound to the Bayes error rate in the two-class case~\cite{fukunaga1990bhatta}. For the special case when the concept scores from both ID and OOD data follow a multivariate Gaussian with a shared covariance matrix, it can be shown that the Bhattacharya distance reduces to the following separability metric (ignoring scale factors): 
\begin{align}
\label{eq:separability_trace}
&J_{\textrm{sep}}(\bfC) ~:=~ J_{\textrm{sep}}(V_{\textrm{in}}(\bfC), V_{\textrm{out}}(\bfC)) ~=~ \textrm{tr}\big[\bfS_w^{-1} \,\bfS_b\big].
%
%
\end{align}

\subsection{Per-class Detection Completeness}
\label{sec:appendix-perclass-completeness}
We propose a per-class measure for the detection completeness (denoted by $\eta^{y}_{\bff, S}(\bfC)$), which is obtained by simply modifying $\eta^{}_{\bff, S}(\bfC)$ in Eqn. (\ref{equ: completeness-detection}) based on the subset of ID and OOD data that are predicted into class $y \in [L]$ by the classifier.

\begin{definition}
\label{def:completeness_detec_perclass}
Given a trained DNN classifier $\bff = \bfh \circ \bfphi$, a trained OOD detector with score function $S(\bfx, \bff)$, and a set of concept vectors $\bfC$, the {\em per-class detection completeness} relative to class $y \in [L]$ with respect to the ID distribution $\Pin(\bfx, y)$ and OOD distribution $\Pout(\bfx)$ is defined as
\begin{align}
\label{equ: completeness-detection-perclass}
    \eta^{y}_{\bff, S}(\bfC) 
    ~:=~ \frac{\textrm{sup}_\bfg \,\textrm{AUC}^y(\bfh \circ \recphi) ~-~ b_r}{\textrm{AUC}(\bfh \circ \bfphi) ~-~ b_r},
\end{align}
\end{definition}
where $\textrm{AUC}^y(\bfh \circ \recphi)$ is the AUROC of the detector conditioned on the event that the class predicted by the concept-world classifier $\,\bfh \circ \recphi\,$ is $y$. We note that the denominator in the above metric is still the global AUROC. The baseline AUROC $b_r$ is equal to $0.5$.
This per-class detection completeness is used in the modified Shapley value defined in Section \ref{sec:expt_concept_based_explanations}.

\subsection{Per-class Concept Separability}
\label{sec:appendix-perclass-separability}
In section \ref{sec:separability_score}, we focused on the separability between the concept scores of ID and OOD data without considering the class prediction of the classifier.
However, it would be more appropriate to impose a high separability between the concept scores on a per-class level. In other words, we would like the concept scores of detected-ID and detected-OOD data, that are predicted by the classifier into any given class $y \in [L]$ to be well separated.
Consider the set of concept-score vectors from the detected-ID (or detected-OOD) dataset that are also predicted into class $y$:
\begin{align}
V^{y}_{\textrm{in}}(\bfC) ~&:=~ \{\TVC(\bfx), ~\bfx \in \Dintr \cup \Douttr ~:~ \calD_\gamma(\bfx, \bff) = 1~ \text{ and } ~\widehat{y}(\bfx) = y\} \nonumber \\
V^{y}_{\textrm{out}}(\bfC) ~&:=~ \{\TVC(\bfx), ~\bfx \in \Dintr \cup \Douttr ~:~ \calD_\gamma(\bfx, \bff) = 0~ \text{ and } ~\widehat{y}(\bfx) = y\}.
\end{align}
We can extend the definition of the global separability metric in Eq. (\ref{eq:separability_trace}) to a given predicted class $y \in [L]$ as follows
\begin{align}
\label{eq:separability_trace_per_class}
J^y_{\textrm{sep}}(\bfC) ~&:=~ J_{\textrm{sep}}(V^{y}_{\textrm{in}}(\bfC), V^{y}_{\textrm{out}}(\bfC)) ~=~ \textrm{tr}\big[(\bfS^{y}_w)^{-1} \,\bfS^y_b\big] \nonumber \\
&=~ (\bfmu^y_{\textrm{out}} \,-\, \bfmu^y_{\textrm{in}})^T \,(\bfS^y_w)^{-1} \,(\bfmu^y_{\textrm{out}} \,-\, \bfmu^y_{\textrm{in}}).
\end{align}
We refer to these per-class variations as \textit{per-class concept separability}.
The scatter matrices $\bfS^y_w$ and $\bfS^y_b$ are defined similar to Eq. (\ref{eq:scatter_matrices}), using the per-class subset of concept scores $V^{y}_{\textrm{in}}(\bfC)$ or $V^{y}_{\textrm{out}}(\bfC)$, and the mean concept-score vectors from the detected-ID and detected-OOD dataset are also defined at a per-class level.


\subsection{Algorithm for Concept Learning}
\label{sec:appendix-algo}
To provide the readers with a clear overview of the proposed concept learning approach, we include Algorithm \ref{alg:concept_learning}.
Note that in line 7 of Algorithm \ref{alg:concept_learning}, the dimension reduction step in $\,V_{\textrm{in}}(\bfC) = \{\TVC(\bfx), ~\bfx \in \Dintr \cup \Douttr ~:~ \calD_\gamma(\bfx, \bff) = 1\}\,$ and $\,V_{\textrm{out}}(\bfC) = \{\TVC(\bfx), ~\bfx \in \Dintr \cup \Douttr ~:~ \calD_\gamma(\bfx, \bff) = 0\}\,$ involves the maximum function, which is not differentiable; specifically, the step $\,\widetilde{v}_{\bfc_i}(\bfx) = \max_{p, q} |\langle \bfphi^{p,q}(\bfx), \bfc_i \rangle|$.
For calculating the gradients (backward pass), we use the \texttt{log-sum-exp} function with a temperature parameter to get a differentiable approximation of the maximum function, \ie $\max_{p, q} |\langle \bfphi^{p,q}(\bfx), \bfc_i \rangle| \,\approx\, \alpha \,\log \left[ \sum_{p,q} \exp\left( \frac{1}{\alpha} \,|\langle \bfphi^{p,q}(\bfx), \bfc_i \rangle| \right) \right]\,$ as $\alpha \rightarrow 0$.
In our experiments, we set the temperature constant $\alpha = 0.001$ upon checking that the approximate value of $\,\widetilde{v}_{\bfc_i}(\bfx)$ is sufficiently close to the original value using the maximum function.

\begin{algorithm}[ht]
\caption{Learning concepts for OOD detector}
\label{alg:concept_learning}
\textbf{INPUT:} Entire training set $\Dtr = \{\Dintr, \Douttr\}$, entire validation set $\Dval = \{\Dinval, \Doutval\}$, classifier $\bff$, detector $\calD_\gamma$. \\
\textbf{INITIALIZE:} Concept vectors $\bfC = [\bfc_1 \cdots \bfc_m]$ and parameters of the network $\bfg$. \\
\textbf{OUTPUT:} $\bfC$ and $\bfg$.
\begin{algorithmic}[1]
  \STATE Calculate threshold $\gamma$ for $\calD_\gamma$ using $\Dval$ as the score at which true positive rate is $95\%$.
  \FOR{$t = 1, ... T$ epochs}
    \STATE Compute the prediction accuracy of the concept-world classifier $\fcon$ using $\Dintr$.
    \STATE Compute the explainability regularization term as defined in \cite{yeh2020completeness}.
	\STATE Compute difference of feature representation between canonical world and concept world (i.e. $J_{\textrm{norm}}(\bfC, \bfg)$).
	\STATE Compute difference of detector outputs between canonical world and concept world using Eqn. (\ref{equ:regularizer_mse}).
	\STATE Compute $V_{\textrm{in}}(\bfC)$ and $V_{\textrm{out}}(\bfC)$ using $\Dtr, \calD_\gamma$ and $\bfC$.
	\STATE Compute separability between $V_{\textrm{in}}(\bfC)$ and $V_{\textrm{out}}(\bfC)$ using Eqn. (\ref{eq:separability_trace}) or Eqn. (\ref{eq:separability_trace_per_class}).
    \STATE Perform a batch-SGD update of $\bfC$ and $g$ using Eqn. (\ref{equ: concept learning}) as the objective.
  \ENDFOR
\end{algorithmic} 
\end{algorithm}

\section{Implementation Details}
\label{sec:appendix-implementation-details}
We ran all our experiments with Tensorflow, Keras and NVDIA GeForce RTX 2080Ti GPUs. We used test-set bootstrapping with 200 runs to obtain the confidence interval for each hyperparameter setting of concept learning.

\subsection{Experimental Setting.}
\mypara{OOD Datasets.}
For the auxiliary OOD dataset for concept learning ($\Douttr$), we use the unlabeled images from MSCOCO dataset (120K images in total) \cite{lin2014mscoco}. We carefully curate the dataset to make sure that no images contain overlapping animal objects with our ID dataset (\ie 50 animal classes of Animals-with-Attributes \cite{xian2018awa}), then randomly sample 30K images.
For OOD datasets for evaluation ($\Doutte$), we use the high-resolution image datasets processed by Huang and Li~\cite{Huang_MOS}.

\mypara{Hyperparameters for Concept Learning.}
Throughout the experiments, we fix the number of concepts to $m = 100$ (unless specifically mentioned otherwise), and following the implementation of \cite{yeh2020completeness}, we set $\lambda_{\textrm{expl}} = 10$ and $\bfg$ to be a two-layer fully-connected neural network with $500$ neurons in the hidden layer.
We learn concepts based on feature representations from the layer right before the global max-pooling layer of the Inception-V3 model.
After concept learning with $m$ concepts, we remove any duplicate (redundant) concept vectors by removing those with a dot product larger than $0.95$ with the remaining concept vectors~\cite{yeh2020completeness}.

\subsection{Additional Results on the Effectiveness of Our Concept Learning}
\label{sec:app_addi_results}

\mypara{Ablation Study for Concept Learning.}
\label{sec:appendix-concept-learning-ablation}
We perform an ablation study that isolates the effect of each regularization term in our concept learning objective (Eqn. \ref{equ: concept learning}) towards our evaluation metrics: classification completeness, detection completeness, and relative concept separability. 
We also observe the coherency among the learned concepts by varying $\lambda_\textrm{mse}$ and $\lambda_\textrm{sep}$.
Coherency of concepts was introduced by Ghorbani \etal~\cite{ghorbani2019ace} to ensure that the generated concept-based explanations are understandable to humans. 
It captures the idea that the examples for a concept should be similar to each other, while being different from the examples corresponding to other concepts.
For the specific case of the image domain, the receptive fields most correlated to a concept $i$ (\eg "stripe pattern") should look different from the receptive fields for a different concept $j$ (\eg "wavy surface of sea").
\citet{yeh2020completeness} proposed to quantify the coherency of concepts as 
\begin{equation}
\label{eq:coherency}
    \frac{1}{m\,K} \mysum_{i=1}^m \mysum_{\bfx^\prime \in T_{\bfc_i}} \langle \bfphi(\bfx^\prime), \bfc_i \rangle,
\end{equation}
where $T_{\bfc_i}$ is the set of $K$-nearest neighbor patches of the concept vector $\bfc_i$ from the ID training set $\Dintr$.

\begin{figure*}[hbt]
  \centering
  \begin{subfigure}{0.45\linewidth}
    \includegraphics[width=\textwidth]{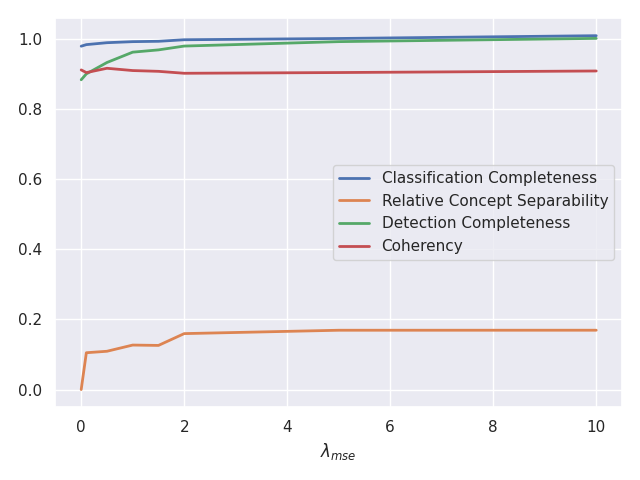}
    \caption{Ablation study varying $\lambda_\textrm{mse}$; we set $\lambda_\textrm{norm} = 0.1, \lambda_\textrm{sep} = 0$}
    \label{fig:ablation_mse}
  \end{subfigure}
  \hfill
  \begin{subfigure}{0.45\linewidth}
    \includegraphics[width=\textwidth]{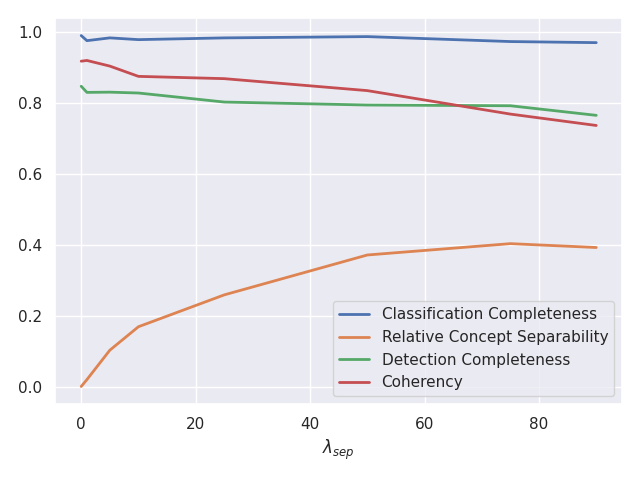}
    \caption{Ablation study varying $\lambda_\textrm{sep}$; we set $\lambda_\textrm{mse} = 0, \lambda_\textrm{norm} = 0$.}
    \label{fig:ablation_sep}
  \end{subfigure}
  \caption{\textbf{Ablation study with respect to $J_\textrm{mse}(\bfC, \bfg)$ and $J_\textrm{sep}(\bfC)$.} We fix $m = 100, \lambda_\textrm{expl} = 10$, and the OOD detector used for concept learing and evaluation is Energy \cite{liu2020energy}}
\label{fig:ablation}
\end{figure*}

We use this metric to quantify how understandable our concepts are for different hyperparameter choices. Figure~\ref{fig:ablation} shows that aligned with our intuition, large $\lambda_\textrm{mse}$ helps to improve the detection completeness. 
Having non-zero $\lambda_\textrm{mse}$ is also helpful to improve the classification completeness even further, and surprisingly concept separability as well, without sacrificing the coherency of concepts.
On the other hand, on the right side of Figure~\ref{fig:ablation}, we observe that large relative concept separability with large $\lambda_\textrm{sep}$ comes at the expense of lower detection completeness and coherency. 
Recall that when visualizing what each concept represents for human's convenience, we apply threshold 0.8 to only presents (see Figure \ref{fig:app-shap}). 
Low coherency with respect to Eqn. \ref{eq:coherency} (\ie 0.768 with $\lambda_\textrm{sep} = 75)$ means that there is much less number of examples that can pass the threshold, meaning that users can hardly understand what the concepts at hand entail.
This observation suggests that one needs to balance between concept coherency and concept separability depending on which property would be more useful for a specific application of concepts.

\mypara{Effectiveness of the Concept Learning.}
In Table~\ref{tab:concept-learning-results}, we present the complete results of concept learning for various combinations of the regularization coefficients across various real-world, large-scale OOD data: \texttt{Places}, \texttt{SUN} and \texttt{Textures}.

\begin{table*}[htb]
    \centering
    \begin{adjustbox}{width=1\textwidth,center}
		\begin{tabular}{l|l|l|c|c|c|c|c|c}
			\toprule
			\multirow{3}{0.001\linewidth}{OOD detector} & \multirow{3}{0.10\linewidth}{Hyper-\\parameters} &
			\multirow{3}{0.05\linewidth}{ $\eta^{}_{\bff}(\bfC) \uparrow$} & \multicolumn{6}{c}{Test OOD dataset} \\ \cline{4-9}
    		& & & \multicolumn{2}{c|}{\texttt{Places}} & \multicolumn{2}{c|}{\texttt{SUN}} & \multicolumn{2}{c}{\texttt{Textures}}\\ \cline{4-9}
    		& & & $\eta^{}_{\bff, S}(\bfC) \uparrow$ & $J_{\textrm{sep}}(\bfC, \bfC') \uparrow$ & $\eta^{}_{\bff, S}(\bfC) \uparrow$ & $J_{\textrm{sep}}(\bfC, \bfC') \uparrow$ & $\eta^{}_{\bff, S}(\bfC) \uparrow$ & $J_{\textrm{sep}}(\bfC, \bfC') \uparrow$ \\ \hline \hline
            \multirow{4}{0.10\linewidth}{MSP} 
			& $(0, 0, 0)$ & 0.977 $\pm$ 0.0006 & 0.774 $\pm$ 0.0010 & 0.694 $\pm$ 0.0153 & 0.782 $\pm$ 0.0010 & 1.088 $\pm$ 0.0175 & 0.593 $\pm$ 0.0013 & 0.765 $\pm$ 0.0157\\
			& $(10, 0.1, 0)$ & \textbf{0.994} $\pm$ 0.0004 & \underline{0.947} $\pm$ 0.0004 & 1.892 $\pm$ 0.0393 & \underline{0.946} $\pm$ 0.0004 & 3.074 $\pm$ 0.0531 & \underline{0.920} $\pm$ 0.0005 & \underline{3.577} $\pm$ 0.1292\\
			& $(0, 0, 50)$ & 0.980 $\pm$ 0.0005 & 0.814 $\pm$ 0.0008 & \underline{2.533} $\pm$ 0.0714 & 0.816 $\pm$ 0.0009 & \underline{4.295} $\pm$ 0.1048 & 0.773 $\pm$ 0.0010 & 3.147 $\pm$ 0.2076\\
			& $(10, 0.1, 50)$ & \underline{0.984} $\pm$ 0.0004 & \textbf{0.960} $\pm$ 0.0004 & \textbf{2.756} $\pm$ 0.0854 & \textbf{0.961} $\pm$ 0.0005 & \textbf{4.442} $\pm$ 0.0830 & \textbf{0.937} $\pm$ 0.0004 & \textbf{3.587} $\pm$ 0.2145\\ \hline
            \multirow{4}{0.10\linewidth}{ODIN} 
			& $(0, 0, 0)$ & 0.977 $\pm$ 0.0006 & 0.742 $\pm$ 0.0011 & 0.444 $\pm$ 0.0119 & 0.745 $\pm$ 0.0010 & 0.710 $\pm$ 0.0156 & 0.618 $\pm$ 0.0013 & 0.501 $\pm$ 0.0121 \\
			& $(10^8, 0.1, 0)$ & \textbf{0.994} $\pm$ 0.0004 & \underline{0.951} $\pm$ 0.0004 & 1.166 $\pm$ 0.0303 & \underline{0.958} $\pm$ 0.0004 & 2.135 $\pm$ 0.0450 & \underline{0.934} $\pm$ 0.0004 & 2.793 $\pm$ 0.0865\\
			& $(0, 0, 50)$ & 0.987 $\pm$ 0.0004 & 0.899 $\pm$ 0.0007 & \underline{1.785} $\pm$ 0.0669 & 0.911 $\pm$ 0.0006 & \underline{3.814} $\pm$ 0.0768 & 0.793 $\pm$ 0.0008 & \underline{3.046} $\pm$ 0.2845\\
			& $(10^8, 0.1, 50)$ & \underline{0.991} $\pm$ 0,0005 & \textbf{0.973} $\pm$ 0.0009 & \textbf{1.813} $\pm$ 0.0268 & \textbf{0.969} $\pm$ 0.0010 & \textbf{4.000} $\pm$ 0.0094 & \textbf{0.945} $\pm$ 0.0006 & \textbf{3.662} $\pm$ 0.1005\\ \hline
            \multirow{4}{0.10\linewidth}{General-ODIN} 
			& $(0, 0, 0)$ & 0.988 $\pm$ 0.0004 & 0.769 $\pm$ 0.0004 & 0.506 $\pm$ 0.0165 & 0.719 $\pm$ 0.0014 & 0.816 $\pm$ 0.0192 & 0.605 $\pm$ 0.0013 & 0.558 $\pm$ 0.1683\\
			& $(10^6, 0.1, 0)$ & \textbf{0.995} $\pm$ 0.0004 & \underline{0.951} $\pm$ 0.0006 & 1.461 $\pm$ 0.0321 & \underline{0.960} $\pm$ 0.0005 & 3.007 $\pm$ 0.0316 & \underline{0.940} $\pm$ 0.0008 & 2.619 $\pm$ 0.1077\\
			& $(0, 0, 50)$ & 0.981 $\pm$ 0.0004 & 0.859 $\pm$ 0.0007 & \underline{1.814} $\pm$ 0.0685 & 0.803 $\pm$ 0.0006 & \underline{4.204} $\pm$ 0.0159 & 0.826 $\pm$ 0.0008 & \textbf{4.014} $\pm$ 0.2246\\
			& $(10^6, 0.1, 50)$ & \underline{0.990} $\pm$ 0.0005 & \textbf{0.971} $\pm$ 0.0010 & \textbf{1.835} $\pm$ 0.0669 & \textbf{0.963}$\pm$ 0.0004 & \textbf{4.287} $\pm$ 0.0284 & \textbf{0.951} $\pm$ 0.0005 & \underline{3.695} $\pm$ 0.1921 \\ \hline
            \multirow{4}{0.10\linewidth}{Energy} 
			& $(0, 0, 0)$ & 0.977 $\pm$ 0.0006 & 0.671 $\pm$ 0.0012 & 0.453 $\pm$ 0.0121 & 0.682 $\pm$ 0.0012 & 0.675 $\pm$ 0.0148 & 0.557 $\pm$ 0.0014 & 0.521 $\pm$ 0.0131\\
			& $(1. 0.1, 0)$ & \textbf{0.993} $\pm$ 0.0005 & \textbf{0.965} $\pm$ 0.0004 & 1.266 $\pm$ 0.0319 & \textbf{0.963} $\pm$ 0.0004 & 2.125 $\pm$ 0.0413 & \textbf{0.960} $\pm$ 0.0003 & 2.648 $\pm$ 0.0596\\
			& $(0, 0, 50)$ & \underline{0.987} $\pm$ 0.0005 & 0.779 $\pm$ 0.0010 & \textbf{1.920} $\pm$ 0.0725 & 0.793 $\pm$ 0.0009 & \textbf{3.659} $\pm$ 0.0659 & 0.767 $\pm$ 0.0010 & \textbf{4.397} $\pm$ 0.2165 \\
			& $(1, 0.1, 50)$ & 0.980 $\pm$ 0.0005 & \underline{0.943} $\pm$ 0.0005 & \underline{1.839} $\pm$ 0.0662 & \underline{0.941} $\pm$ 0.0005 & \underline{3.421} $\pm$ 0.0619 & \underline{0.936} $\pm$ 0.0005 & \underline{3.917} $\pm$ 0.1691 \\ \hline
			\multirow{4}{0.10\linewidth}{Mahala-\\nobis} 
			& $(0, 0, 0)$ & 0.990 $\pm$ 0.0007 & 0.715 $\pm$ 0.0011 & 0.571 $\pm$ 0.0110 & 0.736 $\pm$ 0.0011 & 0.822 $\pm$ 0.0165 & 0.591 $\pm$ 0.0011 & 0.564 $\pm$ 0.0203 \\
			& $(0.1, 0.1, 0)$ & \textbf{0.994} $\pm$ 0.0004 & \underline{0.950} $\pm$ 0.0009 & 1.532 $\pm$ 0.0351 & \underline{0.960} $\pm$ 0.0010 & 2.276 $\pm$ 0.0466 & \underline{0.938} $\pm$ 0.0004 & 2.915 $\pm$ 0.1132\\
			& $(0, 0, 50)$ & 0.985 $\pm$ 0.0004 & 0.880 $\pm$ 0.0005 & \underline{2.550} $\pm$ 0.0681 & 0.883 $\pm$ 0.0006 & \underline{4.091} $\pm$ 0.1013 & 0.774 $\pm$ 0.0007 & \underline{4.274} $\pm$ 0.2305\\
			& $(0.1, 0.1, 50)$ & \underline{0.992} $\pm$ 0.0006 & \textbf{0.961} $\pm$ 0.0005 & \textbf{2.616} $\pm$ 0.0857 & \textbf{0.966} $\pm$ 0.0005 & \textbf{4.325} $\pm$ 0.0055 & \textbf{0.949} $\pm$ 0.0003 & \textbf{4.308} $\pm$ 0.2011 \\ \bottomrule
		\end{tabular}
	\end{adjustbox}
	\caption[]{
	\small \textbf{Results of concept learning with different parameter settings across various OOD detectors and test OOD datasets.} 
	Hyperparameters are in the order of $(\lambda_\textrm{mse}, \lambda_\textrm{norm}, \lambda_\textrm{sep})$.
	Across the rows (for a given OOD detector and OOD dataset), the best value is \textbf{boldfaced}, and second best value is \underline{underscored}.
	The $95\%$ confidence intervals are estimated by bootstrapping the test set over $200$ trials.}
 \label{tab:concept-learning-results}
\end{table*}

\begin{figure*}
  \centering
  \begin{subfigure}{0.32\linewidth}
    \includegraphics[width=\textwidth]{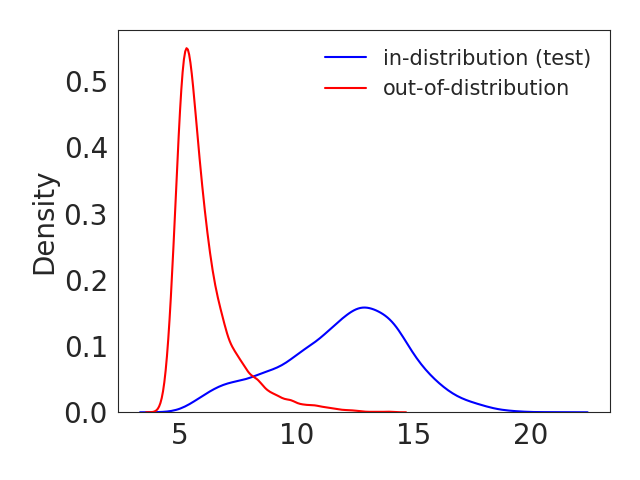}
    \caption{\small Empirical distribution of $S(\bfx, \bff)$ from the target detector.}
    \label{fig:short-a1}
  \end{subfigure}
  \hfill
  \begin{subfigure}{0.32\linewidth}
    \includegraphics[width=\textwidth]{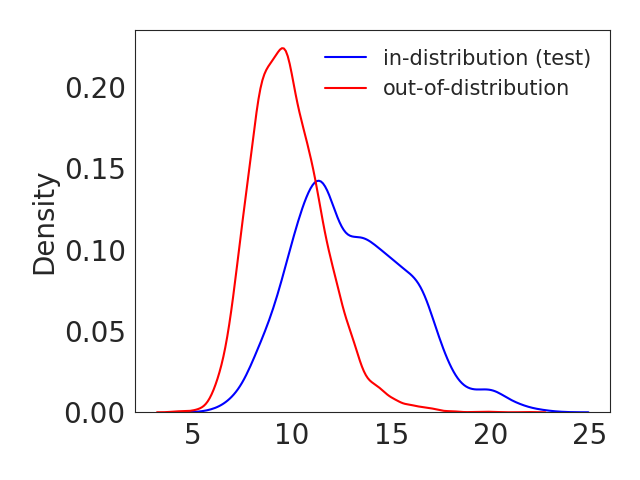}
    \caption{\small Distribution of $\Scon(\bfx, \bff)$ using concepts learned by \citet{yeh2020completeness}.}
    \label{fig:short-b1}
  \end{subfigure}
  \hfill
  \begin{subfigure}{0.32\linewidth}
    \includegraphics[width=\textwidth]{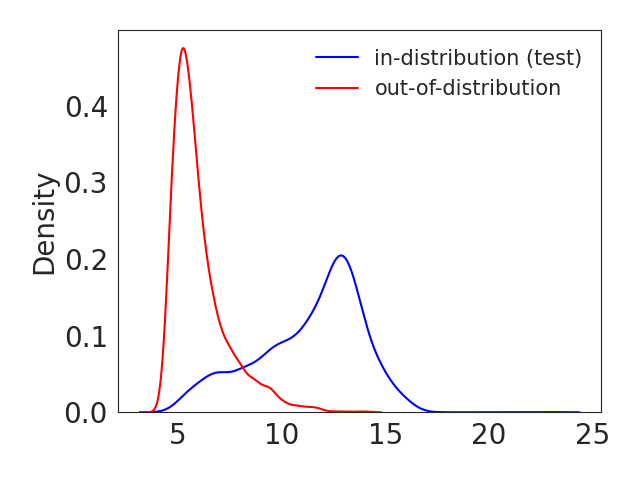}
    \caption{\small Distribution of $\Scon(\bfx, \bff)$ using concepts learned by our method.}
    \label{fig:short-c1}
  \end{subfigure}
  \caption{
  \small (a) Energy detector score $S(\bfx, \bff)$ in the canonical world vs. (b, c) reconstructed $\Scon(\bfx, \bff)$ in the concept world, using different set of concepts.
  Concepts by~\citet{yeh2020completeness} have $\eta^{}_{\bff} = 0.977, ~\eta^{}_{\bff, S}(\bfC) = 0.682$, while concepts by ours $(\lambda_\textrm{mse} = 1, \lambda_\textrm{norm} = 0.1, \lambda_\textrm{sep} = 50)$ have $\eta^{}_{\bff} = 0.984, ~\eta^{}_{\bff, S}(\bfC) = 0.941$.
    Comparison is made between AwA test set (ID, blue) vs. \texttt{SUN} (OOD, red).
    }
\label{fig:score-distribution-energy}
\end{figure*}

\mypara{Accurate Reconstruction of OOD Scores}
In addition to Fig.~\ref{fig:score-distribution-msp}, where we compared the reconstruction accuracy of OOD scores using concepts by \citet{yeh2020completeness} and ours, Fig.~\ref{fig:score-distribution-energy} confirms that the same observation also applies to the Energy detector.

\subsection{Accurate Reconstruction of Classifier Outputs}
\label{app:hellinger}
\begin{figure*}[t]
  \centering
  \includegraphics[width=0.5\textwidth]{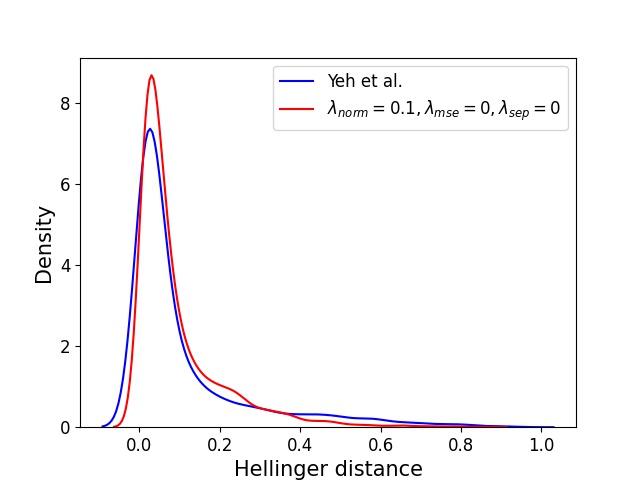}
\caption{Examples for correct detection}
\label{fig:hellinger}
\end{figure*}

We have performed additional experiments to understand if the proposed method can provide improvements in the classification setting. 
Let $\mathbf{C}_1$ denote the concept matrix learned by the method of \citet{yeh2020completeness}. 
Let $\mathbf{C}_2$ denote the concept matrix learned by our method with $\lambda_{mse} = \lambda_{sep} = 0$ and $\lambda_{norm} = 0.1$ (set based on the scale of the regularization term $J_{norm}$). The idea is that we exclude the terms in the concept-learning objective (Eqn.~\ref{equ: concept learning}) that depend on the OOD detector, but include the $\ell_2$ norm based reconstruction error of the layer representation. 
To evaluate the utility of these two sets of concepts for classification, we calculated the per-sample Hellinger distance between the predicted class probabilities of the original classifier and the concept-world classifier (based on either $\mathbf{C}_1$ or $\mathbf{C}_2$). 
Fig.~\ref{fig:hellinger} compares the empirical distribution of the Hellinger distance for both sets of concepts $\mathbf{C}_1$ and $\mathbf{C}_2$. We observe that the distribution is more skewed towards zero with a higher density near zero and a shorter (right) tail in the case of $\mathbf{C}_2$ (red curve) compared to $\mathbf{C}_1$ (blue curve). This suggests that the class predictions are more accurately reconstructed by the concepts learned using our method with only the reconstruction error-based regularization. This can in-turn benefit the concept-based explanations for the classifier.

\section{Choice of Auxiliary OOD Dataset in Concept Learning}
\label{app:auxiliary-ood}

\begin{figure*}[hbt]
\centering
{\includegraphics[width=0.2\textwidth]{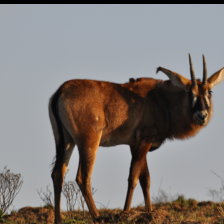}} \hspace{2mm}
{\includegraphics[width=0.2\textwidth]{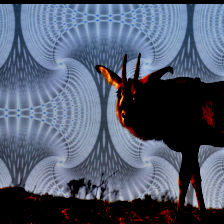}}
\caption{\textbf{Random example of augmented AwA dataset.} 
\textbf{Left:} original image in AwA train set.
\textbf{Right:} corresponding image augmented using the method of \citet{hendrycks2022pixmix}.} 
\label{fig:app-augAwA}
\end{figure*}

Under circumstances where having access to auxiliary OOD dataset for concept learning is not feasible, we suggest that one could use generative methods to generate synthetic dataset, or apply data augmentation techniques. Fig.~\ref{fig:app-augAwA} shows an example of AwA image augmented by \citet{hendrycks2022pixmix}. 

We evaluate the effectiveness of our concept learning objective when such augmented AwA train set is used as auxiliary OOD dataset.
Table~\ref{tab:app-auxiliary-ood} illustrates that the generated concepts with augmented AwA (\ie OOD data close to target ID data) have comparable detection completeness and concept separability compared to when MSCOCO (\ie OOD data far from ID data) was used.
But still, further evaluation on generated concept-based explanations with different choice of auxiliary OOD dataset remains as an interesting research question.

\begin{table}[htb]
    \centering
    \begin{adjustbox}{width=1\columnwidth,center}
		\begin{tabular}{l|l|l|c|c|c|c|c|c}
			\toprule
			\multirow{3}{0.001\linewidth}{OOD detector} & \multirow{3}{0.10\linewidth}{Hyper-\\parameters} &
			\multirow{3}{0.05\linewidth}{ $\eta^{}_{\bff}(\bfC) \uparrow$} & \multicolumn{6}{c}{Test OOD dataset} \\ \cline{4-9}
    		& & & \multicolumn{2}{c|}{\texttt{Places}} & \multicolumn{2}{c|}{\texttt{SUN}} & \multicolumn{2}{c}{\texttt{Textures}}\\ \cline{4-9}
    		& & & $\eta^{}_{\bff, S}(\bfC) \uparrow$ & $J_{\textrm{sep}}(\bfC, \bfC') \uparrow$ & $\eta^{}_{\bff, S}(\bfC) \uparrow$ & $J_{\textrm{sep}}(\bfC, \bfC') \uparrow$ & $\eta^{}_{\bff, S}(\bfC) \uparrow$ & $J_{\textrm{sep}}(\bfC, \bfC') \uparrow$ \\ \hline \hline
			%

            {Energy} 
			& $(1, 0.1, 50)$ & 0.955 $\pm$ 0.0006 & 0.940 $\pm$ 0.0005 & 1.746 $\pm$ 0.0712 & 0.9410 $\pm$ 0.0005 & 3.0703 $\pm$ 0.0580 & 0.927 $\pm$ 0.0005 & 3.417 $\pm$ 0.1419 \\
			\bottomrule
		\end{tabular}
	\end{adjustbox}
        \vspace{2mm}
	\caption[]{Results of concept learning with augmented AwA train set as auxiliary OOD in concept learning.
	}
	\label{tab:app-auxiliary-ood}
\end{table}

\section{Explanations}
\label{sec:appendix-shapleys}

\subsection{Important Concepts for Each OOD Detector}
\label{sec:appendix-explanation}
\begin{figure*}[ht]
\centering
\includegraphics[width=\textwidth]{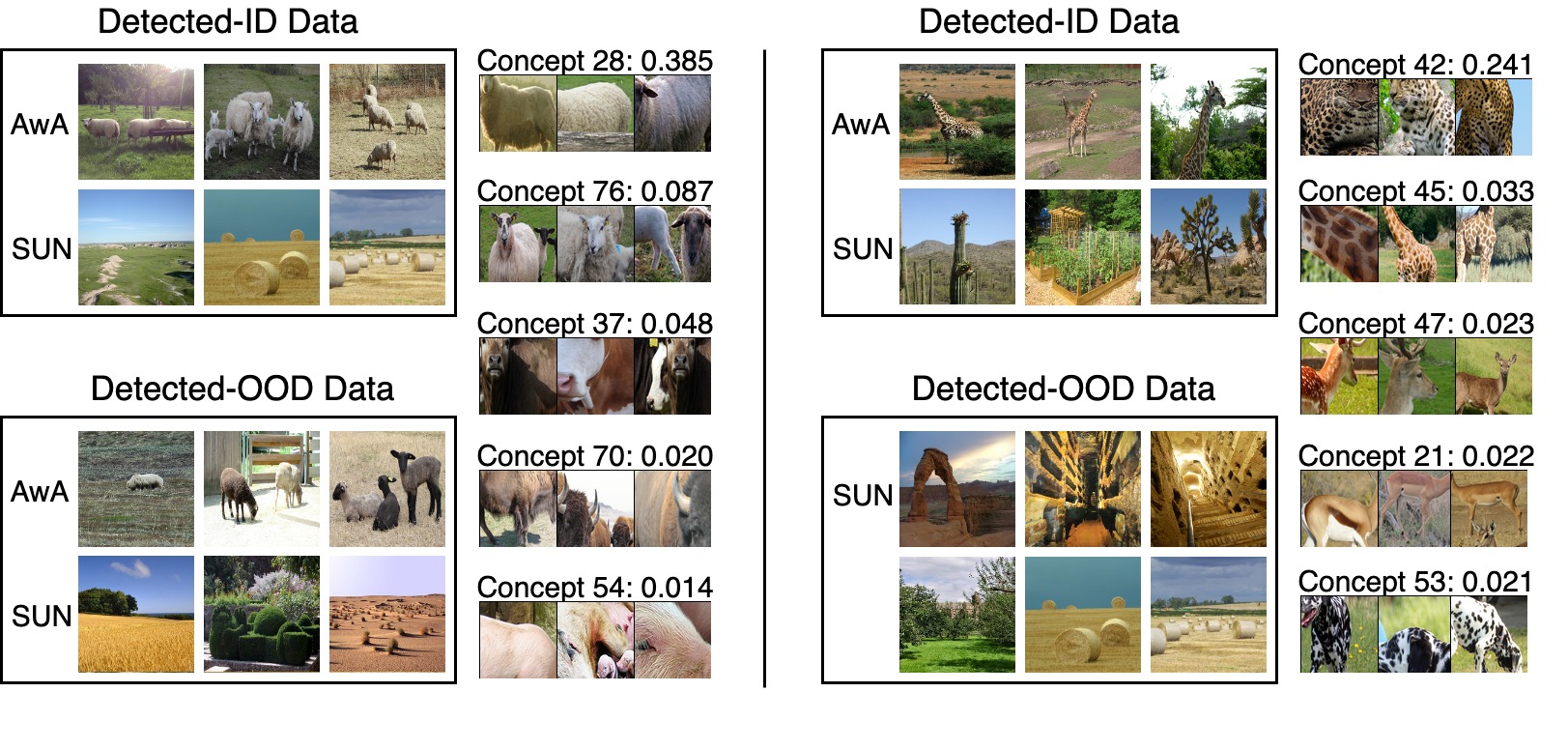}
\vspace{-0.3in}
\caption{Top-6 important concepts for the Energy OOD detector with respect to class ``Sheep'' (on the left) and class ``Giraffe'' (on the right).}
\label{fig:app-shap}
\end{figure*}

We show additional examples for the top-ranked concepts by $\textrm{SHAP}(\eta_{\bff, S}, \bfc_i)$ in Fig. \ref{fig:app-shap}.
For each figure with a fixed choice of class prediction, we present receptive fields from ID test set corresponding to top concepts that contribute the most to the decisions of each OOD detector.
All receptive fields passed the threshold test that the inner product between the feature representation and the corresponding concept vector is over $0.85$.

\begin{figure}[hbt]
\centering
\includegraphics[width=\textwidth]{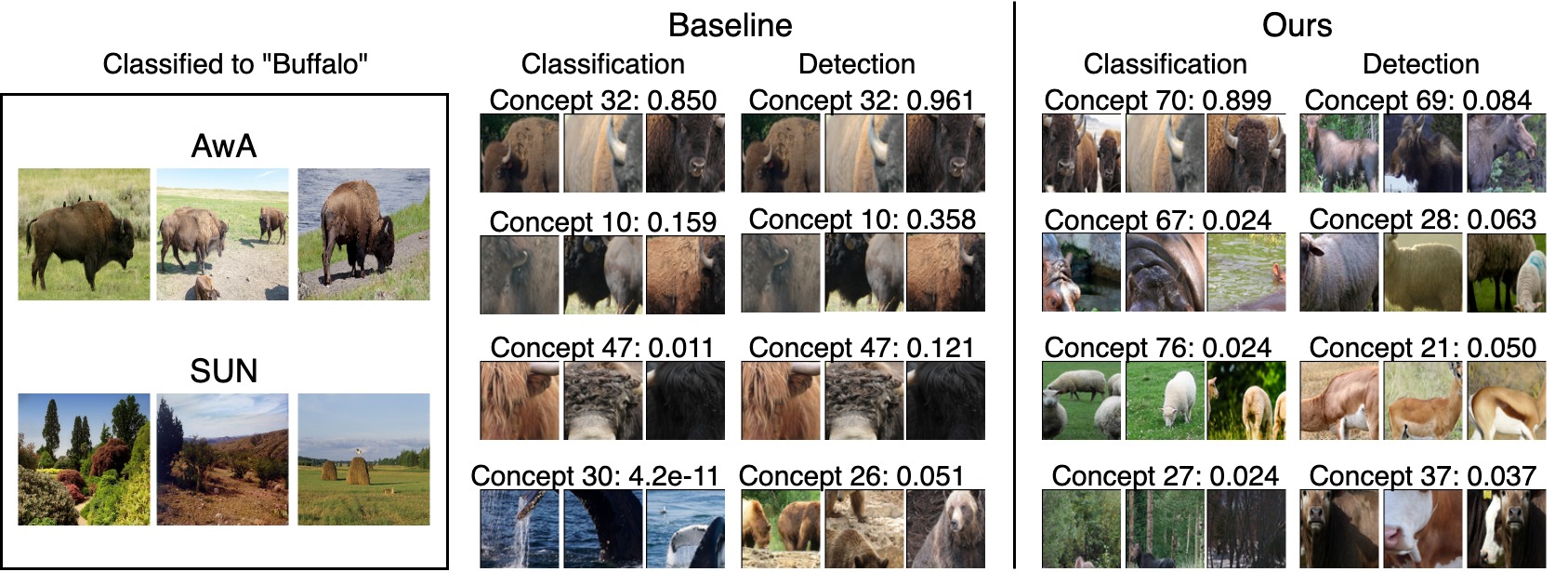}
\caption{\textbf{Most important concepts for the Energy detector with respect to the predicted class ``Buffalo''.} 
We demonstrate randomly sampled images that are predicted by the classifier into this class. 
We compare the top-4 important concepts to describe the DNN classifier (and Energy detector), ranked by the Shapley value based on classification completeness $\textrm{SHAP}(\eta^{j}_{\bff}, \bfc_i)$ (and detection completeness $\textrm{SHAP}(\eta^{j}_{\bff, S}, \bfc_i)$).
``Baseline'' corresponds to the case when the concepts are learned with $\lambda_\textrm{mse} = \lambda_\textrm{norm} = \lambda_\textrm{sep} = 0$, whereas ``Ours'' corresponds to the concepts learned with $\lambda_\textrm{mse} = 1, \lambda_\textrm{norm} = 0.1, \lambda_\textrm{sep} = 0$.
}
\label{fig:shap_buffalo}
\end{figure}

Moreover, in Fig. \ref{fig:shap_buffalo}, we compare the important concepts discovered by the baseline method \cite{yeh2020completeness} (denoted as ``baseline'') vs. ours.
With the baseline, when the learned concepts are solely intended for reconstructing the behavior of the classifier, we observe that interpretation of both the classifier and OOD detector depends on a common set of concepts (\ie concepts 32, 10, and 47).
On the other hand, the concepts learned by our method focus on reconstructing the behavior of both the OOD detector and the classifier. In this case, we observe that a distinct set of important concepts are selected for classification and OOD detection.
We also observe that our method requires more concepts in order to address the decisions of both the classifier and OOD detector.
For instance, the number of concepts obtained by our method and the baseline are 78 and 53 (respectively), out of a total 100 concepts after the duplicate removal of concept vectors.
In short, when the concepts are only targeted at explaining the DNN classifier (as in the baseline \cite{yeh2020completeness}), the behavior of the OOD detector is merely described by the common set of concepts that are important for the DNN classifier.
On the other hand, when not only the DNN classifier but also the OOD detector is taken into consideration during concept learning (\ie our method), we obtain a more diverse and expanded set of concepts, and different concepts play a major role in interpreting the classification and detection results. 

\subsection{More Examples of Our Concept-Based Explanation}
\label{app:more-expl}
In Fig.~\ref{fig:expl-additional1}, we provide additional examples of the concept-based explanations provided by  our method and compare it with that of \cite{yeh2020completeness}.


\begin{figure*}[b]
  \centering
  \begin{subfigure}{\linewidth}
    \includegraphics[width=\textwidth]{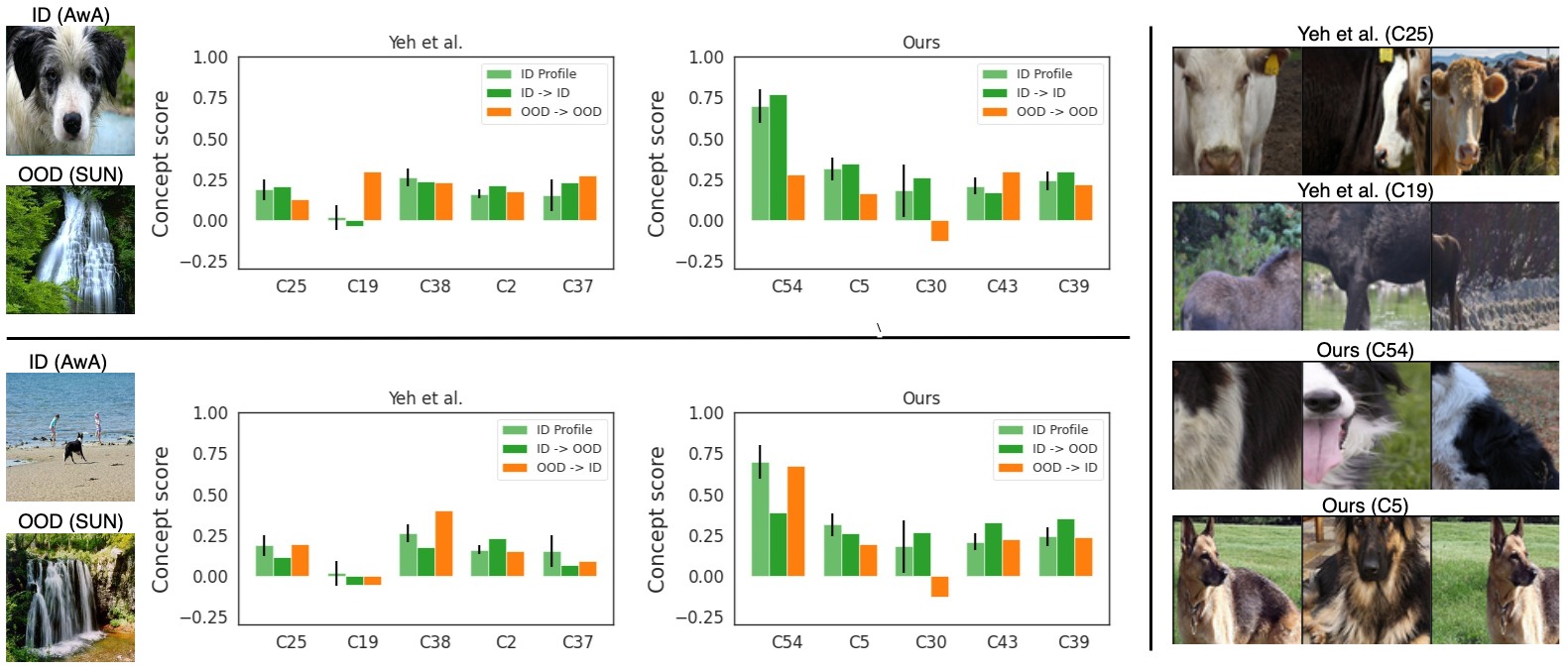}
    \caption{class ``Collie'', Energy OOD detector. Images randomly selected from AwA test set and \texttt{SUN}.}
  \end{subfigure}
  \\
  \begin{subfigure}{\linewidth}
    \includegraphics[width=\textwidth]{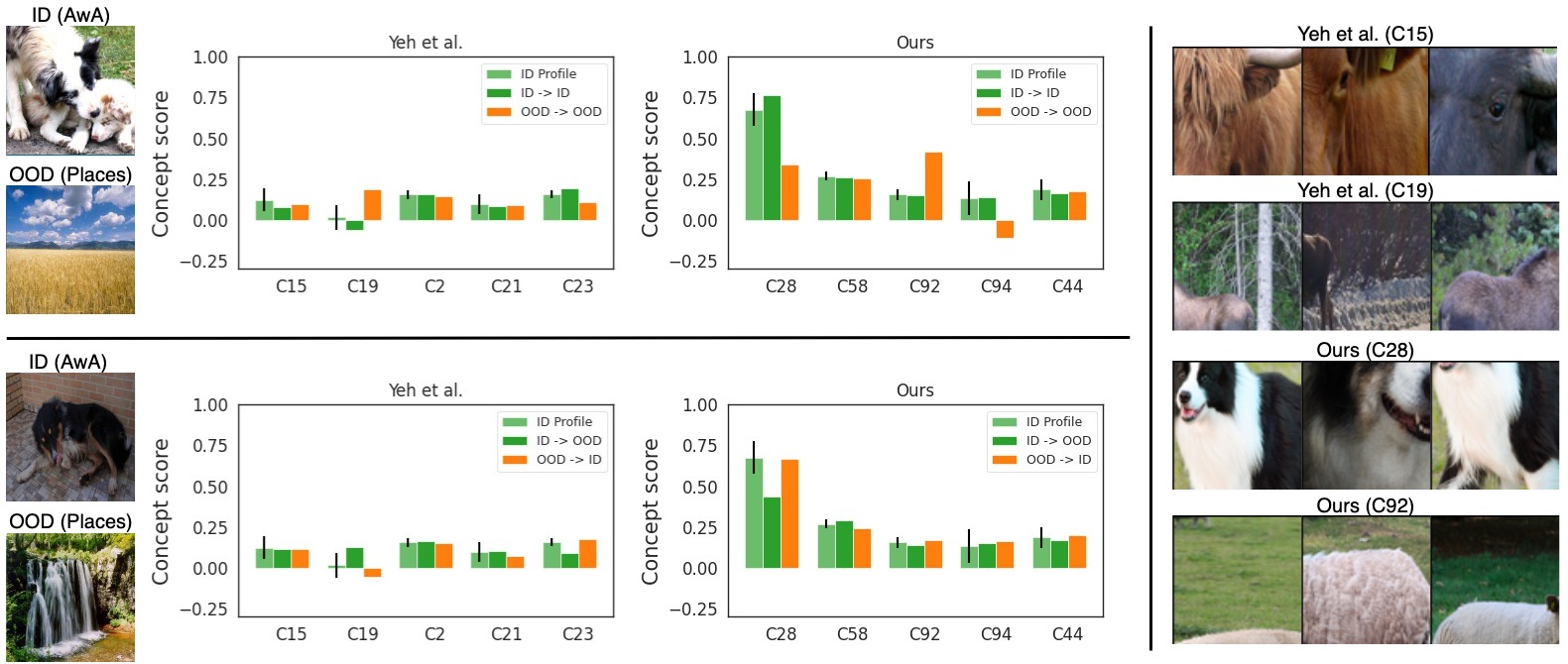}
    \caption{class ``Collie'', MSP OOD detector. Images randomly selected from AwA test set and \texttt{SUN}.}
  \end{subfigure}
\label{fig:expl-additional}
\end{figure*}
  
\begin{figure*}\ContinuedFloat
  \centering
  \begin{subfigure}{\linewidth}
    \includegraphics[width=\textwidth]{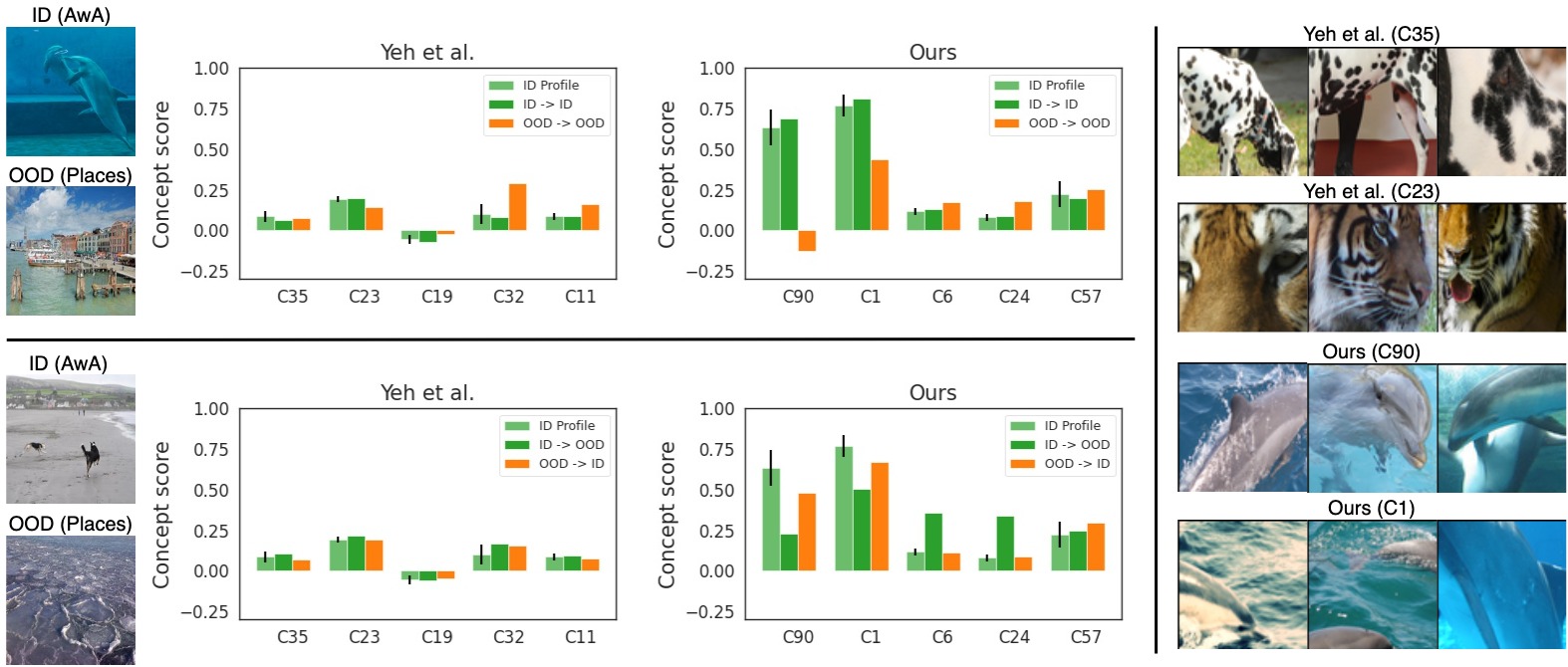}
    \caption{class ``Dolphin'', Energy OOD detector. Images randomly selected from AwA test set and \texttt{Places}.}
    \label{fig:expl-energy-dolphin}
  \end{subfigure}
  \\
  \begin{subfigure}{\linewidth}
    \includegraphics[width=\textwidth]{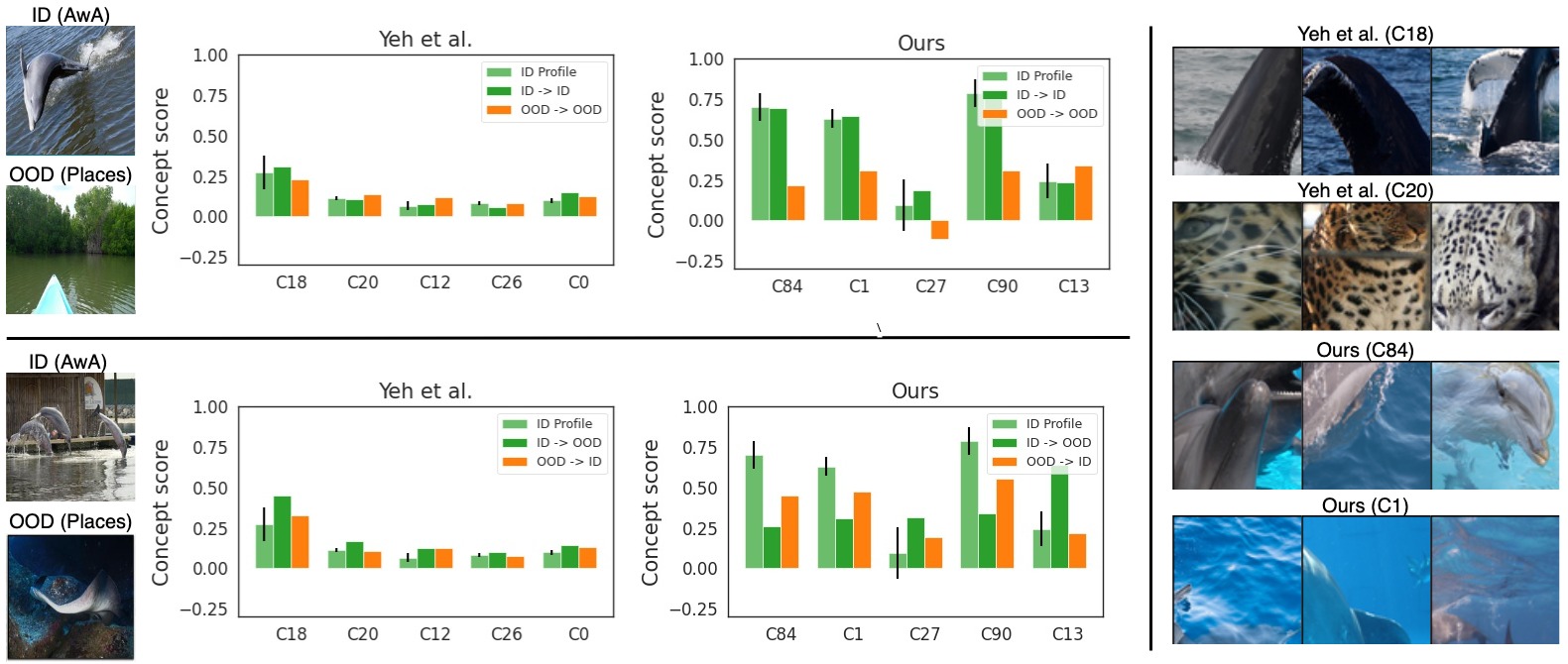}
    \caption{class ``Dolphin'', MSP OOD detector. Images randomly selected from AwA test set and \texttt{Places}.}
    \label{fig:expl-msp-dolphin}
  \end{subfigure}
  \vspace{.1in}
  \caption{\textbf{Concept-based explanations using concepts identified by \citet{yeh2020completeness} vs. ours.}
ID profile shows the average concept-score pattern for normal ID images.}
\label{fig:expl-additional1}
\end{figure*}

\end{document}